\documentclass[10pt,twocolumn,letterpaper]{article}

\usepackage[pagenumbers]{cvpr} 

\makeatletter
\patchcmd{\@maketitle}
  {\vskip .375in}
  {\vskip -.1in}
  {}{}
\makeatother

\usepackage{amsthm}
\newtheorem{proposition}{Proposition}

\usepackage{booktabs}
\usepackage{graphicx}
\usepackage[table]{xcolor}

\definecolor{cvprblue}{rgb}{0.21,0.49,0.74}
\usepackage[pagebackref,breaklinks,colorlinks,allcolors=cvprblue]{hyperref}

\def\confName{CVPR}
\def\confYear{2027}

\title{Propose \& Verify: Fast Reflection Symmetry Detection via Harmonic Fields}

\author{Yishai Lavi\\
Tel Aviv University\\
{\tt\small yishailavi@mail.tau.ac.il}
\and
Shai Avidan\\
Tel Aviv University\\
{\tt\small avidan@eng.tau.ac.il}
}

\begin{document}
\maketitle
\begin{abstract}
Detecting multiple reflection symmetries in 3D shapes
requires aggregating geometric evidence over a large
space of candidate planes, making traditional voting
approaches computationally expensive.
We introduce the Fourier Accelerated Symmetry Transform
(FAST), a training-free framework that combines geometric
pair voting with efficient spectral computation.
Exploiting the fact that global reflection planes pass through
the shape centroid, we reduce the search to plane orientations,
while geometric consistency filtering and independent witnesses
suppress unreliable pair hypotheses.
Our key insight is to represent these votes as a signal over plane orientations and efficiently compute a continuous, smoothed symmetry field in the spherical-harmonic domain.
Specifically, we exploit the Funk-Hecke theorem, the spherical
counterpart of the Fourier convolution theorem for rotationally
symmetric kernels, to transform Gaussian smoothing into
multiplication of harmonic coefficients.
This enables efficient field evaluation,
from which symmetry candidates are extracted, refined, and
geometrically verified.
Across two benchmarks, including a new benchmark of highly
symmetric shapes, FAST achieves the highest clean multi-plane
detection $F_1$ scores among evaluated methods (0.861 and
0.947), with GPU inference below 4 ms per shape: over
$100\times$ faster than the next-fastest evaluated multi-plane GPU baseline.
FAST also demonstrates strong robustness to geometric noise
and competitive single-plane detection, achieving 94.9\%
success on clean single-plane shapes. Our code is publicly available
\href{https://github.com/yishailavi11/propose_and_verify}{here}.
\end{abstract}
\vspace{-0.8em}
    
\begin{figure}[t]
    \centering
    \includegraphics[width=\columnwidth]{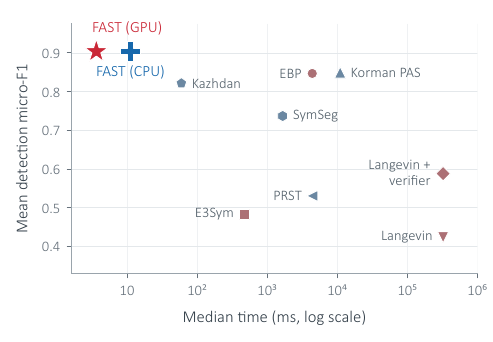} \caption{\textbf{Speed--accuracy comparison on PRS and ANALYTIC.}
    Points show equally weighted means of dataset micro-F1 scores
    and median runtimes (log scale). Blue denotes CPU implementations,
    and red denotes GPU implementations. FAST achieves the highest
    mean accuracy with millisecond inference.}
    \label{fig:speed_accuracy}
    \vspace{-0.8em}
\end{figure}

\section{Introduction}
\label{sec:introduction}

Reflection symmetry is a fundamental property of 3D shapes,
with applications in shape recognition and
reconstruction~\cite{Mitra2013SymmetrySurvey}.
Many objects exhibit multiple reflection planes, making
the recovery of all global symmetries an important geometric
task. However, detecting these planes accurately and
efficiently remains challenging, particularly for noisy
shapes and objects with many symmetries.

Existing approaches include geometric voting~\cite{Podolak:2006:APS},
spectral methods~\cite{Kazhdan:2004:SDA,Martinet:2006:ADS},
and learning-based detectors~\cite{Li_2023_ICCV}.
Pair-voting methods infer candidate planes from surface-point
pairs. Podolak et al.~\cite{Podolak:2006:APS} accumulate evidence
over plane orientation and offset, whereas Je et
al.~\cite{Je2024Langevin} construct a continuous density and
locate its modes through iterative Riemannian Langevin dynamics.
Spectral methods instead analyze global shape representations
or moments. The key observation is that geometric pair votes themselves
define a signal over candidate reflection planes.
Rather than relying on dense accumulation or iterative mode
seeking, this signal can be processed spectrally to construct
and efficiently evaluate a continuous symmetry field.

We introduce the Fourier Accelerated Symmetry Transform (FAST),
a training-free, purely geometric propose-and-verify framework.
Since every exact global reflection plane passes through the
shape centroid, we restrict the search to unoriented plane
normals on the real projective plane $\mathbb{RP}^2$, eliminating
the need to search over plane offsets.
A centroid-consistency gate and independent geometric witnesses
suppress unreliable pair hypotheses before accumulation.

Our central insight is to treat the resulting geometric votes
as a signal over plane orientations, enabling efficient
processing through spherical harmonics.
We construct a continuous symmetry field by smoothing this
signal with a projective geodesic Gaussian kernel.
Crucially, this kernel is zonal, meaning that it depends only
on the relative orientation of two plane normals.
This allows us to exploit the Funk-Hecke theorem
~\cite{muller1966spherical}, the spherical counterpart of
the Fourier convolution theorem for zonal kernels.
Just as the convolution theorem enables efficient filtering
through FFTs, Funk-Hecke transforms our Gaussian smoothing
operation into simple multiplication of harmonic coefficients.
Combined with fast spherical-harmonic transforms and synthesis,
this enables efficient construction and evaluation of the
symmetry field across densely sampled orientations.

For spherical-harmonic bandlimit $L$ and $N_u$ queried directions,
this reduces field construction and evaluation from
$O(N_uL^2)$ with direct evaluation to $O(L^3+N_u)$ at fixed
numerical accuracy. Crucially, this decouples spectral bandwidth
from the density of queried directions, enabling dense
orientation search at modest bandlimits.
We extract candidate planes from local maxima of the field,
continuously refine them, and verify them against the input
geometry.

We evaluate FAST on PRS~\cite{Gao2021PRSNet} and introduce
\textsc{Analytic}, a new benchmark of highly symmetric shapes
with $3$--$15$ exact reflection planes.
FAST achieves the highest clean multi-plane $F_1$ scores
among evaluated methods on both datasets ($0.861$ and $0.947$),
with GPU runtimes below $4$\,ms per shape -- over $100\times$
faster than the next-fastest evaluated multi-plane GPU baseline -- and
CPU runtimes below $13$\,ms (\cref{fig:speed_accuracy}).
It maintains competitive detection accuracy under
geometric noise and achieves 94.9\% clean single-plane
retrieval.

Our main contributions are:
\begin{itemize}
    \item A training-free geometric voting formulation on
    $\mathbb{RP}^2$, using centroid-consistency screening and
    independent witness weighting to suppress unreliable
    hypotheses before accumulation.

    \item A continuous harmonic symmetry field of geometric
    pair votes, exploiting Funk-Hecke convolution and fast
    spherical-harmonic synthesis for efficient, dense
    orientation search.

    \item A fast propose-and-verify detector combining peak
    extraction, continuous refinement, and geometric
    verification, achieving leading clean multi-plane
    detection accuracy and strong performance under noise.

    \item \textsc{Analytic}, a new benchmark with exact
    ground-truth symmetries, enabling systematic evaluation
    of multi-plane detection on highly symmetric 3D shapes.
\end{itemize}

\begin{figure*}[t]
    \centering
    \includegraphics[width=\textwidth]{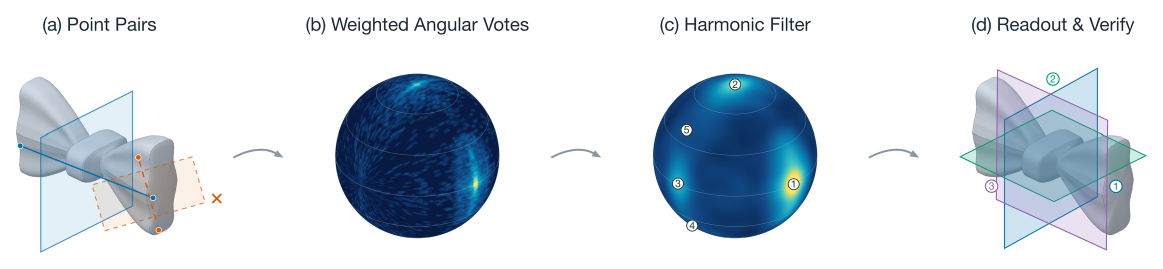}
    \caption{\textbf{Overview of FAST.}
    \textbf{(a)} Each pair of surface points induces a candidate
    reflection plane. The blue
    hypothesis is retained because its plane passes near the
    centroid-centered origin; the orange hypothesis is rejected because its plane is too far from the origin.
    \textbf{(b)} Retained hypotheses cast weighted orientation
    votes on the projective sphere $\mathbb{RP}^2$, where antipodal
    normals represent the same plane.
    \textbf{(c)} Efficient spherical-harmonic filtering produces
    a smooth symmetry field, whose local maxima define candidate
    plane orientations.
    \textbf{(d)} Continuous refinement and geometric verification
    recover three reflection planes (1--3) and reject the
    remaining candidates (4--5).}
    \label{fig:method_overview}
\end{figure*}
\section{Related Work}
\label{sec:related_work}

\paragraph{2D Reflection Symmetry Detection.}
Image-based approaches recover symmetry axes from edge and
orientation evidence~\cite{Reisfeld1995GST,SunSi1999},
local feature matches~\cite{LoyEklundh2006}, and Fourier
or convolutional analysis~\cite{KellerShkolnisky2006,Cicconet2017}.
Other methods use kernel-density voting~\cite{Elawady2017KDE}
or equivariant learning~\cite{Seo_2022_CVPR,Yu_2025_ICCV}.
These methods operate on images rather than 3D geometry.

\paragraph{Geometric 3D Symmetry Detection.}
Early work developed exact symmetry algorithms and continuous
measures of approximate symmetry across
dimensions~\cite{Wolter1985Symmetry,Zabrodsky1995Continuous}.
Geometric approaches include bilateral alignment and
symmetry fitting~\cite{Combes2008Plane,Ecins2017SymFitting},
view-based descriptors~\cite{Li2016View}, and
geometric optimization~\cite{NagarRaman2019Manifold,NagarRaman2020_3DSymm}.
Other methods employ hierarchical
tests~\cite{Zalik2022Hierarchical} or symmetry-guided
segmentation (SymSeg)~\cite{Ecins2018IROS}.
Voting approaches use Hough and Lie-algebra
representations~\cite{Cailliere2008Hough,Shi2016Lie}
or correspondences in partial meshes~\cite{Sipiran2014Partial}.
Korman et al.~\cite{Korman2015PAS} sample transformations
with probabilistic approximation guarantees.
Hruda et al.~\cite{Hruda2020Plane} employ binary witness
tests, while their later work~\cite{Hruda2022Differentiable}
optimizes a continuous symmetry measure.
FAST instead weights individual pair hypotheses using
continuous witness compatibility before accumulation.

Closest to our approach, Mitra et al.~\cite{Mitra2006Partial}
cluster correspondence-induced transformations to recover
partial and approximate symmetries.
Podolak et al.~\cite{Podolak:2006:APS} accumulate point-pair
votes over plane orientation and offset, while Je et
al.~\cite{Je2024Langevin} construct a smoothed density on
the reflection manifold and extract its modes using
Riemannian Langevin dynamics.
FAST instead filters pair hypotheses before accumulating
angular votes and exploits harmonic convolution for
efficient field evaluation, avoiding offset-space
accumulation and iterative mode seeking.

\paragraph{Spectral and Moment-Based 3D Symmetry Detection.}
Spectral approaches analyze global shape representations,
including PCA and extended Gaussian image
correlations~\cite{SunSherrah1997EGI}.
Related methods employ pseudopolar Fourier
analysis~\cite{Bermanis2010Pseudopolar},
spectral self-alignment~\cite{ChertokKeller2010Spectral},
or spherical-harmonic phase~\cite{Kakarala_2013_CVPR}.
Martinet et al.~\cite{Martinet:2006:ADS} recover multiple
symmetries using generalized moments.
Kazhdan et al.~\cite{Kazhdan:2003:ARS,Kazhdan:2004:SDA}
construct continuous spherical symmetry descriptors from
global shape self-similarity using Fourier and harmonic
computation.
FAST instead applies harmonic convolution to geometric
point-pair votes, combining surface-based evidence with
fast arbitrary-direction synthesis for precise multi-plane
localization.

\paragraph{Learning-Based and Feature-Driven 3D Symmetry Detection.}
Neural methods estimate 3D symmetry from learned shape
representations~\cite{JiLiu2019SymNet}
or single-view image
evidence~\cite{Shi2020SymmetryNet,Zhou_2021_CVPR,Li_2025_CVPR}.
PRS-Net~\cite{Gao2021PRSNet} predicts up to three reflection
planes from voxelized shapes, while E3Sym~\cite{Li_2023_ICCV}
clusters learned invariant correspondences to recover
multiple planes.
Aguirre and Sipiran~\cite{Aguirre_2026_WACV}
use pretrained vision features back-projected from rendered
views to detect 3D symmetries.
In contrast, FAST requires neither training nor pretrained
features, deriving symmetry evidence entirely from geometry.

\section{Method}
\label{sec:method}

FAST follows a propose-and-verify strategy.
Surface-point pairs generate candidate reflection planes,
which are screened for centroid consistency and weighted
using independent geometric witnesses.
The retained votes define an angular distribution that is
smoothed and efficiently evaluated as a continuous symmetry
field through harmonic convolution.
Local maxima of this field provide symmetry candidates,
which are refined and geometrically verified.

\subsection{Geometric Pair Voting}
\label{sec:reflection_space}

A reflection plane is represented by a unit normal
$\mathbf n\in\mathbb S^2$ and signed offset $l\in\mathbb R$,
\[
\pi(\mathbf n,l)
=
\{\mathbf z\in\mathbb R^3:
\mathbf n^\top\mathbf z=l\},
\]
where $(\mathbf n,l)$ and $(-\mathbf n,-l)$ represent
the same plane.

Every exact global reflection plane passes through the shape
centroid. After centering the input, we therefore restrict
the search to unoriented plane normals,
\begin{equation}
    \mathcal M
    =
    \mathbb{RP}^2
    =
    \mathbb S^2/\{\mathbf n\sim-\mathbf n\}.
    \label{eq:projective_space}
\end{equation}
We denote an unoriented normal by $[\mathbf n]$.
This reduces the reflection search space to two angular
dimensions, without requiring discretization of plane offsets.

Following geometric pair-voting
methods~\cite{Podolak:2006:APS,Mitra2006Partial,Je2024Langevin},
two distinct surface points $\mathbf p,\mathbf q$ induce
a candidate reflection plane through their perpendicular
bisector,
\begin{equation}
    \mathbf n
    =
    \frac{\mathbf p-\mathbf q}
         {\|\mathbf p-\mathbf q\|_2},
    \qquad
    l
    =
    \frac{1}{2}(\mathbf p+\mathbf q)^\top\mathbf n.
    \label{eq:pair_plane}
\end{equation}
Not every pair is consistent with a global reflection.
We first discard hypotheses whose bisector lies too far
from the centroid, $|l|>\delta$,
where $\delta$ allows deviations due to noise and
imperfect correspondences.
This screening is performed before vote accumulation.

Any two distinct points define a reflection plane that
exchanges them, but this alone does not imply symmetry of
the underlying shape. We therefore assess each retained
hypothesis using $W$ independently sampled witness points
$\{\mathbf z_k\}_{k=1}^{W}$, assigning greater weight to
planes supported by additional geometric evidence.
For a candidate plane $\pi=(\mathbf n,l)$, reflection of
a witness is given by
\begin{equation}
    \mathcal R_\pi(\mathbf z)
    =
    \mathbf z-2(\mathbf n^\top\mathbf z-l)\mathbf n.
\end{equation}
Let $e(\mathbf z)_{\pi}$ be the distance from its reflection
to the observed surface. We define the compatibility and
resulting vote weight, denoted $\widetilde w$, as
\begin{equation}
    c(\mathbf z)_{\pi}
    =
    \exp\left(
        -\frac{e(\mathbf z)_{\pi}^2}{2\tau_w^2}
    \right), \quad
    \widetilde w
    =
    \left(
        \frac{1}{W}\sum_{k=1}^{W}
        c(\mathbf z_k)_{\pi}
    \right)^2,
\label{eq:witness_weight}
\end{equation}
where $\tau_w$ controls the witness tolerance.
Unlike binary consistency tests, this assigns greater
influence to hypotheses supported by independent geometric
evidence. Witnesses are evaluated against the original
pair-induced plane, including its offset; the resulting
weighted vote is accumulated only at its unoriented normal.

\subsection{Continuous Symmetry Field}
\label{sec:continuous_field}

The retained hypotheses define a weighted distribution of
candidate reflection orientations. Votes corresponding to
the same underlying symmetry need not coincide exactly,
particularly under sampling noise. We therefore aggregate
nearby orientations into a smooth field whose peaks indicate
likely reflection planes.

Let $h(\mathbf v)$ denote the normalized, even vote density
on $\mathbb S^2$, satisfying
$h(\mathbf v)=h(-\mathbf v)$.
It represents the weighted vote distribution on
$\mathbb{RP}^2$, where antipodal normals correspond to
the same reflection plane.

The intrinsic distance between two unoriented normals is
\begin{equation}
    d_{\mathcal M}([\mathbf u],[\mathbf v])
    =
    \arccos\left(|\mathbf u^\top\mathbf v|\right).
    \label{eq:projective_geodesic}
\end{equation}
Using this distance, we define a projective geodesic
Gaussian kernel,
\begin{equation}
    K_\sigma([\mathbf u],[\mathbf v])
    =
    \exp\left(
        -\frac{
            d_{\mathcal M}([\mathbf u],[\mathbf v])^2
        }{\sigma^2}
    \right),
    \label{eq:reflection_kernel}
\end{equation}
where $\sigma$ controls the angular smoothing bandwidth.
The continuous symmetry field is then
\begin{equation}
    F(\mathbf u)
    =
    \int_{\mathbb S^2}
    K_\sigma([\mathbf u],[\mathbf v])
    h(\mathbf v)\,d\mathbf v.
    \label{eq:symmetry_field}
\end{equation}
Here, $h$ is the unit-mass even lift of the projective
vote distribution, so no additional normalization factor
is needed. In particular, $F(\mathbf u)=F(-\mathbf u)$.

Direct evaluation of Eq.~\eqref{eq:symmetry_field}
requires integrating over the vote distribution for each
query orientation. Our acceleration exploits the zonal
structure of the projective Gaussian kernel: its dependence
on two plane orientations is determined entirely by their
relative angle. Specifically, since the projective distance
depends only on $|\mathbf u^\top\mathbf v|$, we can write
\begin{equation}
\begin{aligned}
    K_\sigma([\mathbf u],[\mathbf v])
    &= \kappa_\sigma(\mathbf u^\top\mathbf v),\\
    \kappa_\sigma(t)
    &= \exp\left(
        -\frac{\arccos^2(|t|)}{\sigma^2}
    \right).
\end{aligned}
\label{eq:zonal_kernel}
\end{equation}
This rotationally invariant structure makes the symmetry
field a zonal convolution on the sphere, which can be
diagonalized in the spherical-harmonic basis via the
Funk-Hecke theorem.

\subsection{Harmonic Evaluation of the Symmetry Field}
\label{sec:harmonic_field}

We evaluate the zonal convolution using spherical harmonics,
the analogue of the Fourier basis on the sphere.
For zonal kernels, the Funk-Hecke theorem ~\cite{muller1966spherical} provides the spherical counterpart
of the Fourier convolution theorem: convolution becomes
multiplication in the harmonic domain, with a single
kernel-dependent multiplier for each harmonic degree.

Let $\{Y_{\ell m}\}$ be an orthonormal spherical-harmonic
basis and define the coefficients of the vote density by
\begin{equation}
    h_{\ell m}
    =
    \int_{\mathbb S^2}
    h(\mathbf v)\,
    \overline{Y_{\ell m}(\mathbf v)}
    \,d\mathbf v.
    \label{eq:vote_sh_coefficients}
\end{equation}
The Gaussian kernel has a degree-dependent spectrum
\begin{equation}
    \lambda_\ell
    =
    2\pi
    \int_{-1}^{1}
    \kappa_\sigma(t)P_\ell(t)\,dt,
    \label{eq:kernel_spectrum}
\end{equation}
where $P_\ell$ is the Legendre polynomial of degree $\ell$.

Applying the Funk--Hecke theorem to our symmetry field
yields the following result:

\begin{proposition}[Harmonic symmetry field]
\label{prop:harmonic_field}
The spherical-harmonic coefficients of the continuous
symmetry field satisfy
\begin{equation}
    F_{\ell m}
    =
    \lambda_\ell h_{\ell m}.
    \label{eq:harmonic_field}
\end{equation}
Thus, for each harmonic degree $\ell$, convolution with
the projective Gaussian reduces to multiplication by
a single scalar $\lambda_\ell$, shared across all orders $m$.
\end{proposition} 
A derivation is provided in the supplementary material. This identity replaces dense Gaussian interactions between
orientations with coefficient-wise multiplication in the
harmonic domain. Moreover, the kernel spectrum is independent
of the input shape and can be precomputed once.

Because both the vote density and the kernel are
antipodally symmetric, their odd-degree coefficients vanish:
$h_{\ell m}=\lambda_\ell=0$ for odd $\ell$.
The continuous field can therefore be reconstructed using
only even harmonic degrees,
\begin{equation}
    F(\mathbf u)
    =
    \sum_{\substack{\ell\geq0\\
                    \ell\ \mathrm{even}}}
    \sum_{m=-\ell}^{\ell}
    F_{\ell m}Y_{\ell m}(\mathbf u).
    \label{eq:inverse_harmonic_field}
\end{equation}
The result is a continuous field over reflection
orientations, obtained by harmonic filtering rather
than repeated evaluation of the Gaussian integral.

\subsection{Practical Realization}
\label{sec:practical_realization}

\paragraph{Vote Accumulation and Harmonic Projection.}
Let $L$ be our chosen bandlimit. We parameterize directions by $\mu=\cos\vartheta\in[-1,1]$
and azimuth $\phi\in[0,2\pi)$.
The accumulation grid consists of $N_\mu=L+2$
Gauss--Legendre latitude nodes and $N_\phi=4L+1$
uniformly spaced azimuths.
Each retained vote is deposited into its four neighboring
grid nodes using cloud-in-cell (CIC)
interpolation~\cite{EastwoodHockney1974}.
The interpolation weights sum to one, preserving vote mass
while reducing discretization artifacts.

Let $H[q,p]$ denote the resulting histogram of deposited
vote masses. We compute its spherical-harmonic coefficients
up to $L$ by exploiting the separability of the
harmonic basis into azimuthal Fourier modes and latitude
functions. Specifically,
\begin{equation}
\begin{aligned}
    A[q,m]
    &=
    \sum_{p=0}^{N_\phi-1}
    H[q,p]\,e^{-im\phi_p},\\
    \widehat H_{\ell m}
    &=
    \sum_{q=0}^{N_\mu-1}
    A[q,m]\,
    \overline{Y_{\ell m}(\mu_q,0)}.
\end{aligned}
\label{eq:histogram_sht}
\end{equation}
The first step uses a batched real FFT along longitude,
while the second contracts the resulting Fourier
coefficients with precomputed associated-Legendre factors
along latitude.

Because the histogram stores deposited masses rather than
sampled density values, no additional quadrature weights
are required. We restrict the harmonic projection to even
degrees, which is equivalent to symmetrizing the votes, without
explicitly depositing their antipodal counterparts.

\paragraph{Spectral Filtering.}
The shape-independent kernel multipliers are precomputed
from Eq.~\eqref{eq:kernel_spectrum} using Gauss--Legendre
quadrature,
\begin{equation}
    \Lambda[\ell]
    =
    2\pi
    \sum_{a=0}^{N_q-1}
    \omega_a\,
    \kappa_\sigma(\xi_a)P_\ell(\xi_a),
    \label{eq:kernel_quadrature}
\end{equation}
where $\xi_a$ and $\omega_a$ are quadrature nodes and weights.
By the Funk--Hecke theorem, filtering requires only
degree-wise multiplication,
\begin{equation}
    \widehat F_{\ell m}
    =
    \Lambda[\ell]\widehat H_{\ell m}.
    \label{eq:field_coefficients}
\end{equation}

\paragraph{Field Evaluation.}
To extract symmetry candidates, we evaluate the continuous
harmonic field on an icosphere, whose quasi-uniform angular
sampling and triangular connectivity support reliable
local-maximum detection.
Unlike the Gauss--Legendre accumulation grid, the icosphere
avoids highly nonuniform sampling near the poles.
Antipodal vertices are identified, yielding $N_u$
projective query directions $\{[\mathbf u_k]\}_{k=1}^{N_u}$.
The band-limited field at these directions is
\begin{equation}
    F[k]
    =
    \sum_{\substack{0\leq\ell\leq L\\
                    \ell\ \mathrm{even}}}
    \sum_{m=-\ell}^{\ell}
    \widehat F_{\ell m}Y_{\ell m}(\mathbf u_k).
    \label{eq:inverse_sht}
\end{equation}
We evaluate this expansion using fast arbitrary-direction
spherical-harmonic synthesis
~\cite{Reinecke2023GeneralSHT,Belkner2024CunuSHT},
avoiding explicit evaluation of the harmonic basis at
every query direction.
Our selected configuration uses $L=48$ and a level-6
icosphere with $N_u=20{,}481$ projective directions.
Kernel multipliers, latitude basis functions, and reusable
transform data are precomputed.
Further implementation details are provided in the
supplementary material.

\begin{table*}[t]
\centering
\small
\setlength{\tabcolsep}{5pt}
\renewcommand{\arraystretch}{1.10}
\begin{tabular*}{\textwidth}{@{\extracolsep{\fill}}lcccccc@{}}
\toprule
& \multicolumn{3}{c}{PRS}
& \multicolumn{3}{c}{\textsc{Analytic}} \\
\cmidrule(lr){2-4} \cmidrule(l){5-7}
Method
& $F_1\uparrow$ & Angle ($^\circ$)$\downarrow$ & Time$\downarrow$
& $F_1\uparrow$ & Angle ($^\circ$)$\downarrow$ & Time$\downarrow$ \\
\midrule
FAST (ours, GPU)
    & \textbf{0.8612} & \textbf{0.191} & \textbf{3.51\,ms}
    & \textbf{0.9466} & \textbf{0.329} & \textbf{3.76\,ms} \\
EBP~\cite{Aguirre_2026_WACV}
    & 0.7680 & 0.294 & 4.28\,s
    & 0.9263 & 0.452 & 4.50\,s \\
E3Sym~\cite{Li_2023_ICCV}
    & 0.6943 & 0.782 & 475.98\,ms
    & 0.2720 & 1.217 & 466.18\,ms \\
Langevin + verifier
    & 0.6531 & 0.274 & 323.07\,s
    & 0.5233 & 0.832 & 324.75\,s \\
Langevin~\cite{Je2024Langevin}
    & 0.3796 & 0.294 & 323.07\,s
    & 0.4680 & 0.860 & 324.74\,s \\
\midrule
FAST (ours, CPU)
    & \textbf{0.8612} & 0.191 & \textbf{10.25\,ms}
    & \textbf{0.9466} & 0.329 & \textbf{12.09\,ms} \\
Korman PAS~\cite{Korman2015PAS}
    & 0.8002 & 0.513 & 4.99\,s
    & 0.8993 & 0.583 & 17.00\,s \\
Kazhdan~\cite{Kazhdan:2004:SDA}
    & 0.7613 & 1.321 & 64.29\,ms
    & 0.8825 & 1.466 & 54.59\,ms \\
SymSeg~\cite{Ecins2018IROS}
    & 0.6386 & \textbf{0.027} & 1.85\,s
    & 0.8357 & \textbf{0.004} & 1.47\,s \\
PRST~\cite{Podolak:2006:APS}
    & 0.6647 & 1.643 & 3.48\,s
    & 0.3966 & 1.634 & 5.30\,s \\
\bottomrule
\end{tabular*}

\caption{\textbf{Clean multi-plane reflection symmetry detection on
PRS and \textsc{Analytic}.} The upper and lower blocks report GPU
(RTX A5000) and CPU (Intel i7-13700K) implementations, respectively.
Bold indicates the best result within each hardware block and dataset.
We report micro-$F_1$, median angular error ($^\circ$) over matched
true positives, and median runtime over the full datasets. Runtimes are measured on each method’s prepared input representation. Where supported, CPU methods use multithreaded
implementations.}
\label{tab:main_clean}
\end{table*}

\subsection{Complexity}

After angular vote deposition, the accumulation grid has
$N_\mu,N_\phi=\Theta(L)$, giving $\Theta(L^2)$ histogram bins.
Direct Gaussian evaluation from this histogram at $N_u$
query directions requires $\mathcal O(N_uL^2)$ operations.

Our harmonic formulation combines separable coefficient
construction, Funk--Hecke filtering, and fast field synthesis.
The longitude FFT requires
$\mathcal O(L^2\log L)$ operations, followed by a
latitude-only contraction over harmonic degrees and orders
in $\mathcal O(L^3)$.
Funk--Hecke filtering applies degree-wise multipliers
in $\mathcal O(L^2)$ operations.
Finally, fast arbitrary-direction spherical-harmonic
synthesis~\cite{Reinecke2023GeneralSHT,Belkner2024CunuSHT}
requires $\mathcal O(L^3+L^2\log L+N_u)$ operations
at fixed numerical accuracy, compared with
$\mathcal O(N_uL^2)$ for multiplication by a
precomputed dense harmonic basis.
Thus, the total complexity of field construction
and evaluation, after angular binning, is
\begin{equation}
    \boxed{\mathcal O(L^3+N_u)}
    \label{eq:field_complexity}
\end{equation}
This excludes reusable precomputation, geometric vote
generation and deposition, and subsequent candidate
extraction, refinement, and verification.

Crucially, spectral bandwidth and query density can be
selected independently.
Our supplementary ablations show that accurate detection requires only
a modest bandlimit, whereas precise angular localization,
particularly under noise, benefits from dense directional
sampling.
FAST therefore operates in the favorable regime of small
$L$ and large $N_u$, where harmonic evaluation substantially
reduces the cost of searching for symmetry peaks.

\subsection{Candidate Readout}
\label{sec:candidate_readout}

We identify local maxima of the synthesized field using
the triangular neighborhoods of the projective icosphere,
accounting for antipodal equivalence.
A relative peak-height threshold removes weak modes,
and greedy non-maximum suppression eliminates nearby
duplicate orientations.

Each surviving candidate is continuously refined in the
band-limited harmonic field using a local tangent-plane
quadratic approximation.
Since a maximum of the voting field need not minimize
geometric reflection error, refinement is followed by
independent verification against the observed shape.

For both the original and refined plane, we reflect a fixed
subset of surface points and compute exact nearest-neighbor
residuals against the observed geometry.
We retain the orientation with the smaller high-percentile
residual and accept it only if the residual falls below a
fixed threshold.
Both hypotheses are constrained to pass through the
shape centroid.
Further details of peak extraction, refinement, and
verification are provided in the supplementary material.

\section{Results}
\label{sec:results}

\subsection{Datasets and Setup}
\label{sec:experimental_setup}

We evaluate global reflection symmetry detection on complete
3D shapes using two complementary benchmarks.
The PRS dataset~\cite{Gao2021PRSNet} comprises $750$ test
shapes with $894$ annotated reflection planes, and a disjoint
validation set of $100$ shapes.
To evaluate shapes with larger symmetry sets, we additionally
introduce the ANALYTIC benchmark, comprising 120 observations
of procedurally constructed polyhedra, including prisms,
pyramids, bipyramids, frusta, unequal-sided cuboids, and
Platonic solids. The benchmark contains 987 analytically
determined reflection planes, with 3--15 planes per
observation. Construction details are provided in the
supplementary material.

For FAST, we uniformly sample $100{,}000$ surface points
per mesh. Competing methods use their respective input
representations, as detailed in the supplementary material.
Reported runtimes begin from each method's prepared input
representation, with CPU implementations using multithreading
where supported.
We evaluate clean inputs and isotropic Gaussian noise at
$1\%$, $3\%$, and $5\%$ RMS displacement relative to the
clean bounding-box diagonal.

FAST hyperparameters are calibrated on the clean PRS
validation set, with the harmonic bandlimit selected based
on the accuracy--efficiency trade-off examined in our
supplemental ablation study. The resulting configuration
is fixed across all test datasets and noise levels.
Where applicable, competing methods are calibrated on
the same set by adjusting their final detection
threshold. All configurations remain fixed throughout
testing.

Predicted and ground-truth planes are matched one-to-one
using Hungarian assignment~\cite{Kuhn1955Hungarian},
accounting for the equivalence
$(\mathbf n,l)\sim(-\mathbf n,-l)$.
A match is considered correct if the angular error
is at most $5^\circ$ and the offset error is at most
$0.02D$, where $D$ is the observed shape's bounding-box
diagonal. Unmatched predictions and annotations count
as false positives and false negatives, respectively.

\begin{figure}[t]
    \centering
    \includegraphics[width=\columnwidth]{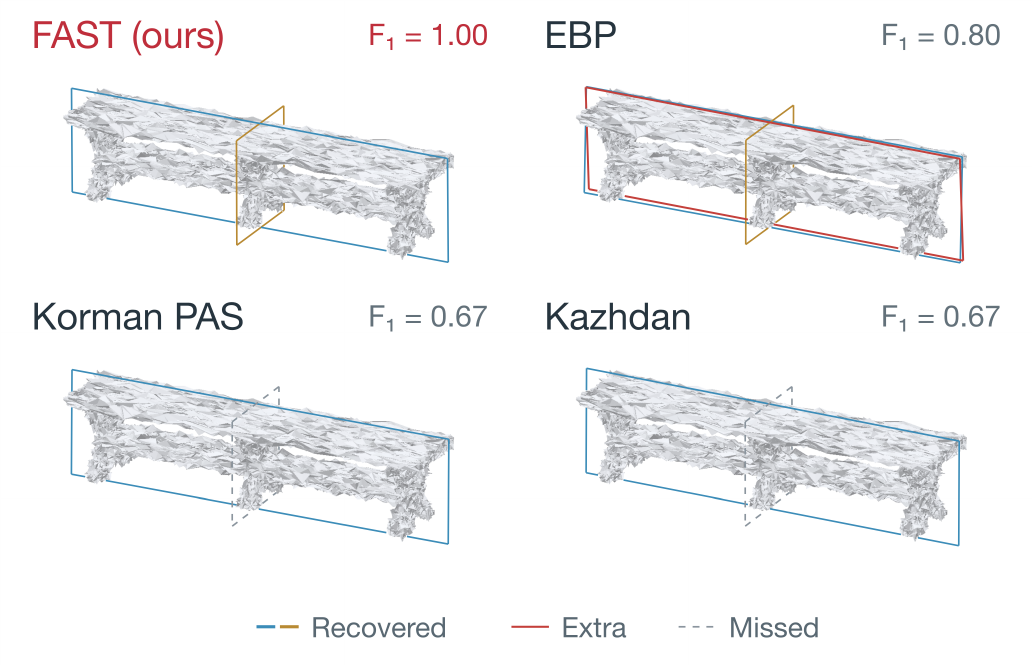}
    \caption{\textbf{Qualitative comparison on a PRS bench with 1\% RMS noise.} FAST recovers both annotated reflection planes without extra detections. 
    }
    \label{fig:qualitative_prs}
\end{figure}

\subsection{Comparison with Existing Methods}
\label{sec:main_comparison}

Table~\ref{tab:main_clean} compares detection accuracy,
angular localization, and runtime on clean shapes across
both benchmarks.
We evaluate multi-plane reflection symmetry detectors with
publicly available implementations.
For Podolak et al.~\cite{Podolak:2006:APS}, we use a corrected
third-party Python reimplementation.
We exclude Mitra et al.~\cite{Mitra2006Partial} and
Martinet et al.~\cite{Martinet:2006:ADS}, for which faithful
implementations were unavailable.
Since Langevin was designed for partial symmetry detection,
we additionally evaluate its proposals using our geometric
verifier to assess their suitability for global symmetry
detection.

FAST achieves the highest clean multi-plane $F_1$ on both
benchmarks, reaching $0.8612$ on PRS and $0.9466$ on
\textsc{Analytic}.
On PRS, it outperforms the strongest competing method,
Korman PAS ($0.8002$), by $6.10$ percentage points.
On \textsc{Analytic}, FAST surpasses EBP ($0.9263$) by
$2.03$ percentage points, while substantially outperforming
classical geometric voting and spectral methods such as
PRST and Kazhdan.
FAST also achieves the lowest median angular errors among
the evaluated GPU methods, at $0.191^\circ$ and
$0.329^\circ$, respectively.

SymSeg~\cite{Ecins2018IROS} achieves the lowest angular
errors overall, at $0.027^\circ$ on PRS and $0.004^\circ$
on \textsc{Analytic}.
Its ICP-like approach iteratively refines reflection planes
using symmetric point correspondences and point-to-plane
geometric fitting~\cite{Ecins2017SymFitting}.
However, it achieves lower detection $F_1$ ($0.639$ and
$0.836$) than FAST, with substantially longer runtimes
of $1.85$ and $1.47$\,s per shape, respectively.

Beyond accuracy, FAST requires only $3.51$ and
$3.76$\,ms per shape on GPU, respectively, making it
approximately $136\times$ and $124\times$ faster than
E3Sym, the next-fastest evaluated GPU baseline.
Compared with EBP, FAST is over $1{,}000\times$ faster
on both benchmarks while achieving higher detection $F_1$.
Our CPU implementation achieves identical detection
accuracy in just $10.25$ and $12.09$\,ms, approximately
$6.3\times$ and $4.5\times$ faster than Kazhdan,
the next-fastest evaluated CPU baseline.
These results establish FAST as an accurate, training-free
multi-plane detector with millisecond-scale inference
on both CPU and GPU.



\begin{figure}[t]
    \centering
    \includegraphics[width=\linewidth]{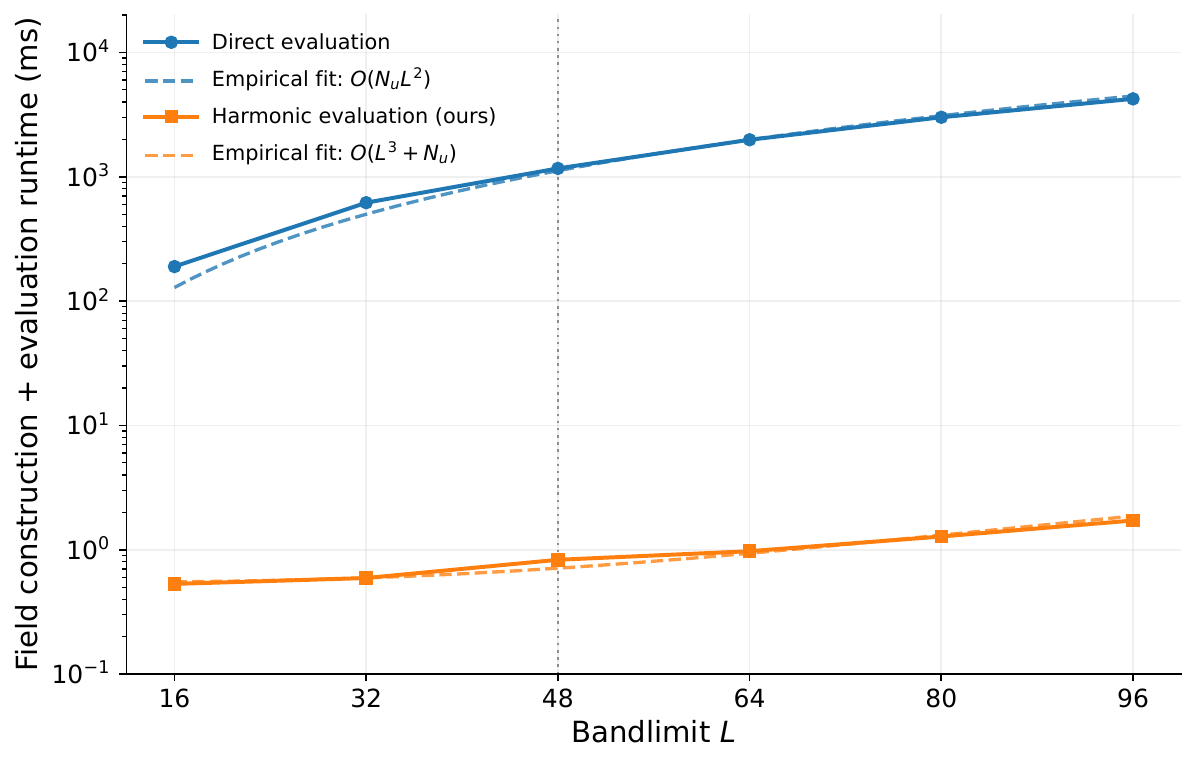}
    \caption{\textbf{Efficiency of harmonic field construction and evaluation.}
    Median single-threaded CPU runtimes for constructing and querying
    the symmetry field.
    Dashed curves show empirical fits to the predicted complexity
    forms.
    At our chosen band limit $L=48$ (dotted line), FAST achieves a $1{,}405\times$ speedup.}
    \label{fig:harmonic_efficiency}
\end{figure}
\subsection{Harmonic Operator Efficiency}
\label{sec:operator_efficiency}

We empirically evaluate the theoretical complexity
reduction of our harmonic formulation. We compare field construction and
evaluation against direct summation of the projective
geodesic Gaussian, using identical normalized vote
histograms and query directions. Both implementations
use vectorized numerical operations on a single CPU thread.
We exclude voting, candidate extraction, refinement,
verification, and reusable precomputation.

We vary $L=16,32,\ldots,96$ and evaluate the field at
20,481 projective directions. At our chosen $L=48$,
direct evaluation takes $1{,}166.873$\,ms, compared with
just $0.830$\,ms for FAST: a $1{,}405\times$ speedup
(Fig.~\ref{fig:harmonic_efficiency}), with pooled relative
field error of only $1.56\times10^{-5}$.

The measured scaling is consistent with our complexity
analysis: direct evaluation requires $O(N_uL^2)$
operations, whereas harmonic construction and fast
synthesis require $O(L^3+N_u)$ at fixed numerical
accuracy, where $N_u$ is the number of query directions.
Empirical scaling curves are fitted across six bandlimits
and three icosphere resolutions, demonstrating how the
theoretical computational advantage translates into
substantial runtime savings in practice.

\begin{table}[t]
\centering
\footnotesize
\setlength{\tabcolsep}{1.5pt}
\renewcommand{\arraystretch}{1.08}
\begin{tabular*}{\columnwidth}{@{\extracolsep{\fill}}lcccccc@{}}
\toprule
& \multicolumn{3}{c}{PRS}
& \multicolumn{3}{c}{\textsc{Analytic}} \\
\cmidrule(lr){2-4} \cmidrule(l){5-7}
Method & 1\% & 3\% & 5\% & 1\% & 3\% & 5\% \\
\midrule
FAST (ours)
& \cellcolor{red!25}\textbf{0.850}
& \cellcolor{red!25}\textbf{0.751}
& \cellcolor{red!25}\textbf{0.647}
& \cellcolor{orange!30}0.875
& \cellcolor{red!25}\textbf{0.764}
& \cellcolor{orange!30}0.642 \\
Korman PAS~\cite{Korman2015PAS}
& \cellcolor{yellow!35}0.760
& 0.179 & 0.002
& \cellcolor{red!25}\textbf{0.887}
& 0.032 & 0.000 \\
Kazhdan~\cite{Kazhdan:2004:SDA}
& \cellcolor{orange!30}0.833
& \cellcolor{orange!30}0.747
& \cellcolor{orange!30}0.644
& \cellcolor{yellow!35}0.806
& \cellcolor{orange!30}0.763
& \cellcolor{red!25}\textbf{0.737} \\
EBP~\cite{Aguirre_2026_WACV}
& 0.639 & 0.416 & 0.236
& 0.426 & 0.016 & 0.000 \\
SymSeg~\cite{Ecins2018IROS}
& 0.033 & 0.000 & 0.000
& 0.649 & 0.000 & 0.000 \\
PRST~\cite{Podolak:2006:APS}
& 0.595 & 0.411 & 0.263
& 0.403 & 0.304
& \cellcolor{yellow!35}0.208 \\
E3Sym~\cite{Li_2023_ICCV}
& 0.655 & \cellcolor{yellow!35}0.527 & 0.377
& 0.193 & 0.075 & 0.044 \\
Langevin + verifier
& 0.611 & 0.519 & \cellcolor{yellow!35}0.444
& 0.444 & \cellcolor{yellow!35}0.317 & 0.190 \\
Langevin~\cite{Je2024Langevin}
& 0.384 & 0.372 & 0.349
& 0.395 & 0.282 & 0.170 \\
\bottomrule
\end{tabular*}
\caption{\textbf{Noise robustness.}
Method configurations are fixed across noise levels,
with acceptance thresholds calibrated on clean PRS validation set
where applicable. Kazhdan uses subdivided meshes with perturbed vertices,
whereas point-cloud methods use noisy surface samples.
Red, orange, and yellow highlight the first, second, and
third best results in each column, respectively.}
\label{tab:noise_robustness}
\end{table}

\subsection{Robustness to Noise}
\label{sec:noise_robustness}

We evaluate robustness at $1\%$, $3\%$, and $5\%$
isotropic Gaussian noise. Each surface point is
independently perturbed by
$\boldsymbol\epsilon\sim
\mathcal N(\mathbf 0,
(kD_{\mathrm{clean}}/100)^2\mathbf I_3/3)$,
where $k$ is the noise percentage and
$D_{\mathrm{clean}}$ is the clean bounding-box diagonal.
Thus, $k\%$ noise corresponds to an RMS point
displacement of $k\%$ of the clean shape extent.
Each noisy observation is independently centered
and normalized, and all detection thresholds remain
fixed at their clean-validation values.

As shown in Table~\ref{tab:noise_robustness},
FAST achieves the highest $F_1$ on PRS across all
noise levels, reaching $0.850$, $0.751$, and $0.647$
at $1\%$, $3\%$, and $5\%$ noise, respectively.
Kazhdan is the closest competitor, achieving
$0.833$, $0.747$, and $0.644$.
Both methods substantially outperform the remaining
baselines under stronger perturbations.

On \textsc{Analytic}, Korman PAS achieves the highest
$F_1$ at $1\%$ noise ($0.887$), narrowly ahead of
FAST ($0.875$), but its performance deteriorates
sharply at higher noise levels.
FAST achieves the highest $F_1$ at $3\%$ noise
($0.764$), marginally ahead of Kazhdan ($0.763$).
At $5\%$, Kazhdan performs best ($0.737$), while
FAST achieves $0.642$, outperforming the remaining methods.

Overall, FAST maintains strong robustness to geometric
noise, leading across all noise levels on PRS and
remaining competitive on \textsc{Analytic}.

\begin{table}[t]
\centering
\footnotesize
\setlength{\tabcolsep}{2pt}
\renewcommand{\arraystretch}{1.07}
\begin{tabular*}{\columnwidth}{@{\extracolsep{\fill}}lccccc@{}}
\toprule
Method & Clean & 1\% & 3\% & 5\% & Time$\downarrow$ \\
\midrule
FAST (ours, GPU)
& \cellcolor{red!25}\textbf{0.949}
& \cellcolor{red!25}\textbf{0.917}
& \cellcolor{red!25}\textbf{0.848}
& \cellcolor{red!25}\textbf{0.753}
& \cellcolor{red!25}\textbf{3.53\,ms} \\
EBP~\cite{Aguirre_2026_WACV}
& \cellcolor{orange!30}0.915
& \cellcolor{orange!30}0.876
& \cellcolor{orange!30}0.805
& \cellcolor{orange!30}0.671
& 4.22\,s \\
E3Sym~\cite{Li_2023_ICCV}
& 0.736 & 0.664 & 0.475 & 0.296
& \cellcolor{yellow!35}477.81\,ms \\
Langevin~\cite{Je2024Langevin}
& \cellcolor{yellow!35}0.810
& \cellcolor{yellow!35}0.769
& \cellcolor{yellow!35}0.714
& \cellcolor{yellow!35}0.651
& 323.07\,s \\
PRS-Net~\cite{Gao2021PRSNet}
& 0.406 & 0.401 & 0.390 & 0.385
& \cellcolor{orange!30}50.61\,ms \\
\midrule
FAST (ours, CPU)
& \cellcolor{red!25}\textbf{0.949}
& \cellcolor{orange!30}0.917
& \cellcolor{orange!30}0.848
& \cellcolor{yellow!35}0.753
& \cellcolor{orange!30}11.77\,ms \\
Korman PAS~\cite{Korman2015PAS}
& 0.894
& \cellcolor{yellow!35}0.910
& 0.813 & 0.689 & 4.79\,s \\
Kazhdan~\cite{Kazhdan:2004:SDA}
& \cellcolor{orange!30}0.903
& \cellcolor{red!25}\textbf{0.943}
& \cellcolor{red!25}\textbf{0.918}
& \cellcolor{red!25}\textbf{0.879}
& \cellcolor{yellow!35}73.61\,ms \\
SymSeg~\cite{Ecins2018IROS}
& 0.843 & 0.705 & 0.489 & 0.473 & 1.88\,s \\
PRST~\cite{Podolak:2006:APS}
& 0.782 & 0.748 & 0.692 & 0.604 & 3.63\,s \\
PCA
& \cellcolor{yellow!35}0.900
& 0.885
& \cellcolor{yellow!35}0.836
& \cellcolor{orange!30}0.768
& \cellcolor{red!25}\textbf{10.16\,ms} \\
Hruda~\cite{Hruda2022Differentiable}
& 0.789 & 0.748 & 0.221 & 0.047 & 196.54\,ms \\
\bottomrule
\end{tabular*}
\caption{\textbf{Single-plane retrieval on the $611$ PRS
shapes with one annotated reflection plane.}
The upper and lower blocks report GPU and CPU implementations, respectively.
Each method selects its lowest-p99 available candidate
without an acceptance threshold, returning at most one
plane. Colors indicate the top three results
within each hardware block and column.}
\label{tab:single_plane}
\end{table}

\subsection{Single-Plane Detection}
\label{sec:single_plane}

Although our method targets recovery of all reflection
planes, we additionally evaluate its performance in
the single-plane setting.
This allows comparison with methods designed to recover
only one or a fixed number of planes, including
Hruda et al.~\cite{Hruda2022Differentiable},
PRS-Net~\cite{Gao2021PRSNet}, and PCA.
We evaluate all methods on the $611$ PRS shapes with
a single annotated reflection plane.
To evaluate candidate retrieval under a common criterion,
we apply our geometric verifier to each method's available
candidates, selecting the plane with the lowest
$99$th-percentile reflected nearest-neighbor residual.
No acceptance threshold is applied, and each method
returns at most one plane.
Runtimes are reported over all $611$ clean shapes.
FAST uses measured complete-detector runtimes;
baseline runtimes combine each shape's archived native
detector time with separately measured geometric
verification time.

As shown in Table~\ref{tab:single_plane}, FAST
achieves the highest success rate on clean shapes
($0.949$), outperforming EBP ($0.915$) and Kazhdan
($0.903$). Among GPU methods, FAST achieves the
highest accuracy across all noise levels, maintaining
success rates of $0.917$, $0.848$, and $0.753$ at $1\%$,
$3\%$, and $5\%$ noise, respectively. Among CPU methods,
Kazhdan performs best under noise, reaching $0.943$,
$0.918$, and $0.879$, while FAST remains competitive.

FAST also achieves the lowest GPU runtime of
$3.53$\,ms, compared with $50.61$\,ms for PRS-Net,
the next-fastest GPU method. On CPU, FAST requires
$11.77$\,ms, slightly above PCA ($10.16$\,ms) and
substantially faster than the remaining CPU methods.
These results demonstrate that the generality of our
approach does not come at the expense of competitive
single-plane detection performance.

\begin{table}[t]
\centering
\footnotesize
\setlength{\tabcolsep}{2pt}
\renewcommand{\arraystretch}{1.07}
\begin{tabular*}{\columnwidth}{@{\extracolsep{\fill}}lcccccc@{}}
\toprule
& \multicolumn{3}{c}{PRS}
& \multicolumn{3}{c}{\textsc{Analytic}} \\
\cmidrule(lr){2-4} \cmidrule(l){5-7}
Variant
& $F_1\uparrow$ & Angle$\downarrow$ & ms$\downarrow$
& $F_1\uparrow$ & Angle$\downarrow$ & ms$\downarrow$ \\
\midrule
Full FAST
& 0.861 & 0.191 & 3.44
& 0.947 & 0.329 & 3.68 \\
No pair screening
& 0.802 & 0.210 & 3.03
& 0.793 & 0.821 & 3.38 \\
Raw CIC (mass)
& 0.724 & 0.871 & 3.84
& 0.857 & 1.232 & 3.96 \\
No refinement
& 0.863 & 0.445 & 2.69
& 0.944 & 0.536 & 2.92 \\
No verification
& 0.351 & 0.209 & 2.41
& 0.664 & 0.336 & 2.47 \\
Lower bandlimit ($L=16$)
& 0.854 & 0.389 & 3.91
& 0.797 & 0.472 & 3.99 \\
Lower icosphere level (4)
& 0.860 & 0.262 & 3.36
& 0.914 & 0.385 & 3.65 \\
\bottomrule
\end{tabular*}

\caption{\textbf{Clean component and resolution ablations
on PRS and \textsc{Analytic}.}
We report micro-$F_1$, median angular error ($^\circ$)
over matched true positives, and median GPU internal
runtime (ms). Full FAST uses $L=48$ and a level-6
icosphere. No pair screening disables both centroid
filtering and witness weighting. Raw CIC extracts peaks
from the unsmoothed histogram, bypassing harmonic
processing and refinement.}
\label{tab:component_ablations}
\end{table}

\subsection{Ablation Study}
\label{sec:ablation_study}

Table~\ref{tab:component_ablations} evaluates the main
components and resolution choices of FAST on both benchmarks.
All variants follow the same evaluation protocol,
with the remaining parameters held fixed.

Geometric screening and verification are essential for
reliable detection. Disabling both centroid-consistency
filtering and witness weighting reduces $F_1$ from
$0.861$ to $0.802$ on PRS and from $0.947$ to $0.793$
on \textsc{Analytic}. Removing geometric verification
causes an even larger drop, to $0.351$ and $0.664$,
respectively, highlighting its role in rejecting
spurious symmetry candidates.

Harmonic processing and refinement play complementary
roles in detection and localization. Extracting peaks directly from the unsmoothed CIC
histogram, bypassing harmonic processing and refinement,
reduces $F_1$ to $0.724$ and $0.857$, respectively.
Refinement primarily improves localization: removing it
increases median angular error from $0.191^\circ$ to
$0.445^\circ$ on PRS and from $0.329^\circ$ to
$0.536^\circ$ on \textsc{Analytic}, with little effect
on detection $F_1$.

Furthermore, reducing the bandlimit to $L=16$ lowers $F_1$ to
$0.854$ on PRS and $0.797$ on \textsc{Analytic}, while
increasing angular errors. Reducing the icosphere resolution
to level 4 similarly degrades localization and lowers
\textsc{Analytic} $F_1$ to $0.914$.
Additional ablations are provided in the supplementary
material.
\section{Conclusion}
\label{sec:conclusion}

We presented FAST, a training-free framework for detecting
multiple global reflection symmetries in 3D shapes.
By treating geometric pair votes as a signal over plane
orientations, FAST exploits the Funk-Hecke theorem to
efficiently construct and evaluate a continuous symmetry
field in the spherical-harmonic domain.
Geometric screening, witness weighting, continuous refinement,
and verification turn this efficient representation into a
reliable symmetry detector.
Across two benchmarks, FAST achieves the highest clean
multi-plane detection accuracy among evaluated methods,
with GPU inference below $4$\,ms per shape, while
demonstrating strong noise robustness and competitive
single-plane performance.
These results establish harmonic voting as an efficient
alternative to costly geometric search and learned
symmetry detection.

\section*{LLM Usage Statement}

OpenAI ChatGPT was used to assist with literature review, technical discussions, coding, experimental design and analysis, dataset construction, figure preparation, and manuscript drafting and editing. Google Deep Review was used for pre-submission manuscript feedback. The authors reviewed and verified all AI-assisted content and take full responsibility for the methods, results, and conclusions presented in this work.

{
    \small
    \bibliographystyle{ieeenat_fullname}
    \bibliography{main}
}

\clearpage
\setcounter{page}{1}
\maketitlesupplementary

\section{Qualitative Results}
\label{sec:supp_qualitative}

\begin{figure*}[p]
    \centering
    \includegraphics[
        width=\textwidth,
        height=0.86\textheight,
        keepaspectratio
    ]{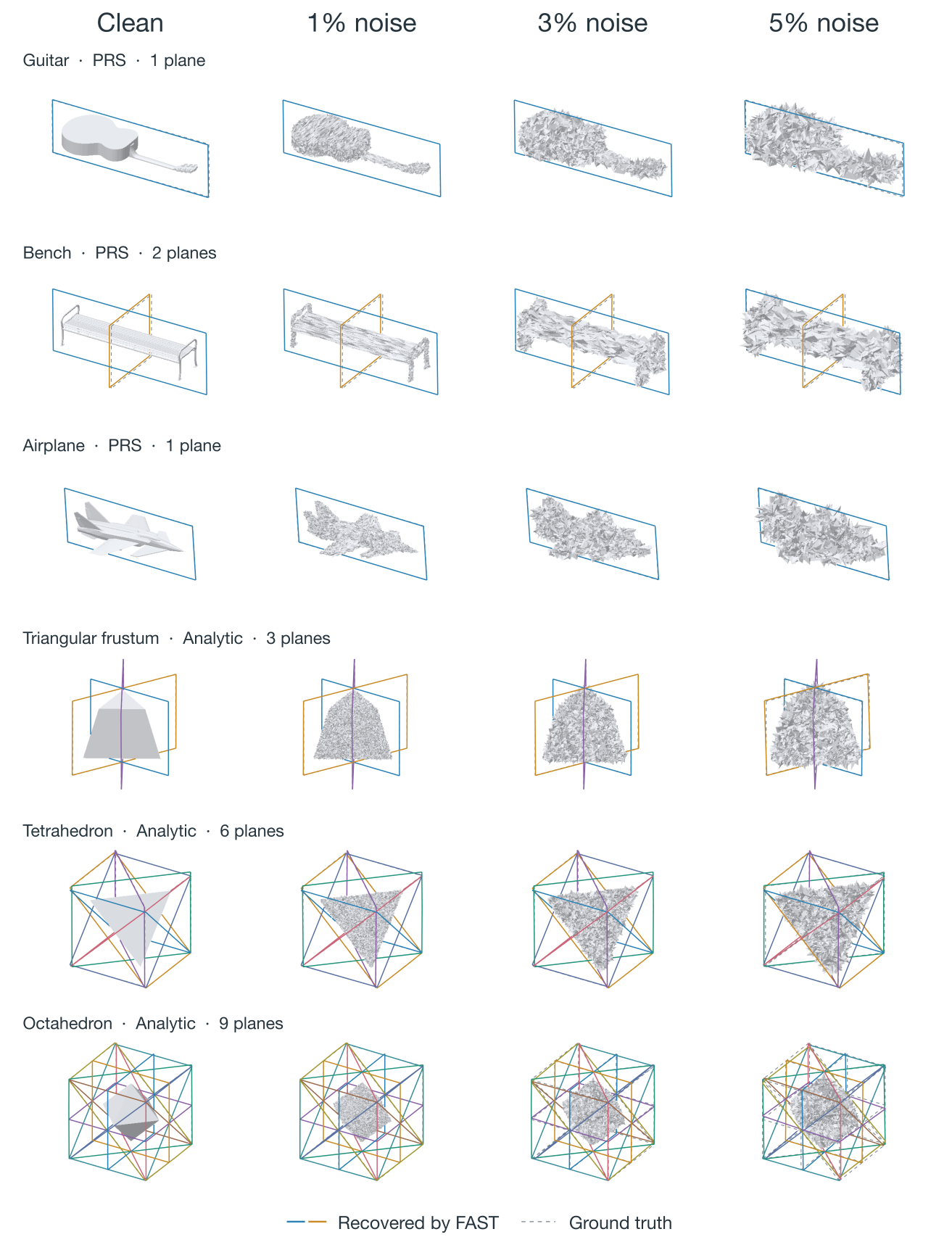}
    \caption{
    \textbf{Qualitative reflection symmetry detection across noise levels.}
    Selected examples from PRS (top three rows) and
    \textsc{Analytic} (bottom three rows), evaluated on clean
    inputs and at $1\%$, $3\%$, and $5\%$ RMS Gaussian noise.
    The examples range from objects with one or two reflection
    planes to highly symmetric polyhedra with three, six, and
    nine planes.
    Colored solid outlines indicate planes recovered by FAST,
    while dashed gray outlines indicate ground-truth planes.
    Noise magnitudes are defined relative to the clean
    bounding-box diagonal.
    }
    \label{fig:supp_qualitative_noise}
\end{figure*}

Figure~\ref{fig:supp_qualitative_noise} illustrates FAST's
behavior across increasing geometric noise and varying
numbers of reflection planes. The PRS examples include
a guitar and an airplane with a single annotated plane,
and a bench with two planes. The \textsc{Analytic}
examples demonstrate recovery of larger symmetry sets,
including three planes for a triangular frustum,
six for a tetrahedron, and nine for an octahedron.

Across these selected examples, the recovered planes
remain closely aligned with the ground-truth annotations
as noise increases from clean observations to $5\%$ RMS.
This is particularly notable for the highly symmetric
polyhedra, where multiple intersecting reflection planes
must be recovered simultaneously despite substantial
surface perturbations.

\section{Mathematical Foundations and Proofs}
\label{sec:supp_proofs}

\subsection{Zonality and Invariance}
\label{sec:supp_invariance}

\paragraph{Projective geodesic and zonality.}
A path on $\mathbb{RP}^2$ from $[\mathbf u]$ to
$[\mathbf v]$ lifts to a spherical path from $\mathbf u$
to either $\mathbf v$ or $-\mathbf v$.
Taking the shorter great-circle arc gives
\begin{equation}
\begin{split}
d_{\mathcal M}([\mathbf u],[\mathbf v])
&=\min\{\arccos(\mathbf u^\top\mathbf v),
          \arccos(-\mathbf u^\top\mathbf v)\}\\
&=\arccos(|\mathbf u^\top\mathbf v|).
\end{split}
\label{eq:supp_projective_distance}
\end{equation}
The corresponding Gaussian kernel is
\begin{equation}
\kappa_\sigma(t)
=\exp\!\left[-\frac{\arccos^2(|t|)}{\sigma^2}\right].
\label{eq:supp_projective_kernel}
\end{equation}
Since $\kappa_\sigma(t)=\kappa_\sigma(-t)$, the kernel
is well-defined on equivalence classes of antipodal
normals. It is zonal, depending only on the inner product.
For any $Q\in SO(3)$,
$(Q\mathbf u)^\top(Q\mathbf v)=\mathbf u^\top\mathbf v$;
hence the projective distance and kernel are invariant
under simultaneous rotation.

\paragraph{Rotation equivariance.}
For an angular density $h$, define
$h_Q(\mathbf v)=h(Q^{-1}\mathbf v)$.
By rotational invariance of the spherical measure and
the kernel, a change of variables gives
\begin{equation}
F_{h_Q}(Q\mathbf u)=F_h(\mathbf u).
\end{equation}
Thus, the continuous smoothing operator is rotation
equivariant.

\subsection{Harmonic Representation}
\label{sec:supp_harmonic_proof}

\paragraph{Proof of Proposition~\ref{prop:harmonic_field}.}
Let $\nu$ be the mass-preserving even lift of the
projective vote measure to $\mathbb S^2$: a vote of
mass $w$ at $[\mathbf v]$ corresponds to masses $w/2$
at $\mathbf v$ and $-\mathbf v$.
The continuous symmetry field is
\begin{equation}
F(\mathbf u)=\int_{\mathbb S^2}
\kappa_\sigma(\mathbf u^\top\mathbf v)\,d\nu(\mathbf v).
\end{equation}
This representation preserves the total vote mass,
so no additional normalization factor is required.

Using complex orthonormal spherical harmonics with
surface-area measure $d\Omega$, define
\begin{equation}
h_{\ell m}
=\int_{\mathbb S^2}
\overline{Y_{\ell m}(\mathbf v)}\,d\nu(\mathbf v).
\end{equation}

Since the kernel is bounded and $\nu$ has finite mass,
Fubini's theorem permits exchanging the integrals:
\begin{align}
F_{\ell m}
&=\int_{\mathbb S^2}F(\mathbf u)
  \overline{Y_{\ell m}(\mathbf u)}\,d\Omega(\mathbf u)
  \nonumber\\
&=\int_{\mathbb S^2}\left[
  \int_{\mathbb S^2}
  \kappa_\sigma(\mathbf u^\top\mathbf v)
  \overline{Y_{\ell m}(\mathbf u)}\,d\Omega(\mathbf u)
  \right]d\nu(\mathbf v).
\end{align}
By the Funk--Hecke theorem~\cite{muller1966spherical},
the inner integral equals
$\lambda_\ell\overline{Y_{\ell m}(\mathbf v)}$, where
\begin{equation}
\lambda_\ell
=2\pi\int_{-1}^{1}\kappa_\sigma(t)P_\ell(t)\,dt.
\label{eq:supp_kernel_spectrum}
\end{equation}
Therefore,
\begin{equation}
F_{\ell m}=\lambda_\ell h_{\ell m},
\end{equation}
establishing the degree-wise diagonalization of the
smoothing operator. The argument applies directly to
finite vote measures without requiring a pointwise
harmonic expansion of atomic votes.\hfill$\square$

\paragraph{Antipodal symmetry and even harmonics.}
Spherical harmonics satisfy
\begin{equation}
Y_{\ell m}(-\mathbf v)=(-1)^\ell Y_{\ell m}(\mathbf v).
\end{equation}
Since $\nu$ is even, $h_{\ell m}=0$ for odd $\ell$.
Likewise, the evenness of $\kappa_\sigma$ and the
parity relation $P_\ell(-t)=(-1)^\ell P_\ell(t)$
imply $\lambda_\ell=0$ for odd $\ell$.

The degree-$L$ approximation of the field is therefore
\begin{equation}
F_L(\mathbf u)
=\sum_{\substack{0\leq\ell\leq L\\\ell\ \mathrm{even}}}
 \sum_{m=-\ell}^{\ell}
 \lambda_\ell h_{\ell m}Y_{\ell m}(\mathbf u).
\label{eq:supp_truncated_field}
\end{equation}
Because only even degrees appear,
$F_L(\mathbf u)=F_L(-\mathbf u)$, so the truncated
field is well-defined on $\mathbb{RP}^2$.
The corresponding discrete histogram construction
is described in Section~\ref{sec:supp_histogram}.

\clearpage

\clearpage

\section{Weighted Voting and Histogram Construction}
\label{sec:implementation_details}
\label{sec:supp_histogram}

We describe how sampled point-pair hypotheses are converted
into a discrete angular vote distribution and subsequently
projected onto spherical harmonics.

\subsection{Point-Pair Hypotheses and Centroid Filtering}
\label{sec:supp_configuration}

Each mesh is represented by $N=100{,}000$ area-uniform
surface samples. We subtract their sample centroid
$\mathbf c$ and divide by the observed bounding-box
diagonal $D$, obtaining normalized points
$\mathbf x_i=(\mathbf p_i-\mathbf c)/D$.

We sample $M=250{,}000$ ordered point pairs with distinct
indices, independently across pair slots. Each pair
induces its perpendicular-bisector plane $(\mathbf n,\ell)$,
as defined in the main paper. We retain only hypotheses
satisfying
\begin{equation}
    |\ell|\leq\delta,\qquad \delta=0.0108,
\end{equation}
where the offset is measured in normalized coordinates.
Rejected hypotheses contribute no voting mass.
Crucially, centroid-consistency filtering is applied
before histogram accumulation, so no offset dimension
is required in the vote distribution.

\subsection{Continuous Witness Weighting}
\label{sec:supp_witnesses}

For each pair, we sample $W=2$ independent witness points,
excluding the pair endpoints. Witnesses are sampled with
replacement and may coincide.

To estimate geometric consistency efficiently, we construct
a $96^3$ occupancy grid over the normalized point-cloud
bounding box. Each occupied voxel is represented by its
center. For a reflected witness $\mathcal R_\pi(\mathbf z)$,
we search the $3\times3\times3$ voxel neighborhood
and approximate its distance to the observed surface
by the distance to the nearest occupied voxel center.
If no occupied voxel is found, the residual is infinite.

For witness residuals $e_k$, the compatibility and
pair weight are
\begin{equation}
c_k=\exp\left(-\frac{e_k^2}{2\tau_w^2}\right),
\qquad
\widetilde w=
\left(\frac{1}{W}\sum_{k=1}^{W}c_k\right)^2,
\label{eq:supp_witness_weight}
\end{equation}
with $\tau_w=0.01$.
Witnesses are reflected using the original pair-induced
plane $(\mathbf n,\ell)$, including its offset.

The retained weights are normalized to unit total mass
before harmonic projection.

\subsection{Projective Histogram and Antipodal Symmetrization}
\label{sec:supp_cic}

We represent the angular votes on a Gauss--Legendre grid
with $N_\mu=L+2=50$ latitude nodes and $N_\phi=4L+1=193$
uniform azimuths, where $L=48$. Directions are
parameterized by $\mu=\cos\vartheta$ and
$\phi\in[0,2\pi)$, with grid directions
\begin{equation}
\mathbf v_{qp}=
\begin{pmatrix}
\sqrt{1-\mu_q^2}\cos\phi_p\\
\sqrt{1-\mu_q^2}\sin\phi_p\\
\mu_q
\end{pmatrix}.
\end{equation}

Since $\mathbf n$ and $-\mathbf n$ represent the same
projective direction, we choose a signed representative
using the pair offset: $\mathbf n$ for $\ell\geq0$
and $-\mathbf n$ otherwise.

Each retained vote is deposited once using cloud-in-cell
(CIC) interpolation~\cite{EastwoodHockney1974}.
Its weight is distributed bilinearly among the four
neighboring grid nodes in $(\mu,\phi)$, with interpolation
weights summing to one. Azimuth is periodic, while
directions beyond the extreme Gauss--Legendre latitudes
are clamped to the nearest latitude ring.
The resulting histogram $H[q,p]$ stores deposited masses.

Importantly, the histogram itself need not be
antipodally symmetric. Define its discrete measure
\begin{equation}
d\nu_H=\sum_{q,p}H[q,p]\delta_{\mathbf v_{qp}}.
\end{equation}
Its antipodally symmetrized counterpart is
\begin{equation}
d\nu_H^{+}=
\frac12\sum_{q,p}H[q,p]
\left(\delta_{\mathbf v_{qp}}+
      \delta_{-\mathbf v_{qp}}\right).
\label{eq:supp_even_measure}
\end{equation}
Since
$Y_{\ell m}(-\mathbf v)=(-1)^\ell Y_{\ell m}(\mathbf v)$,
the even-degree harmonic coefficients of $d\nu_H$
and $d\nu_H^{+}$ are identical, while the odd-degree
coefficients of the latter vanish.

Thus, restricting harmonic projection to even degrees
is equivalent to symmetrizing the deposited vote measure,
without explicitly depositing a second antipodal vote.
The resulting field satisfies
$F_L(\mathbf u)=F_L(-\mathbf u)$.

\subsection{Harmonic Projection and Field Synthesis}
\label{sec:supp_mass_projection}

Because $H[q,p]$ contains deposited masses, its
spherical-harmonic coefficients are obtained by
discrete mass projection:
\begin{equation}
\widehat H_{\ell m}
=\sum_{q,p}H[q,p]\,
\overline{Y_{\ell m}(\mathbf v_{qp})},
\qquad \ell\ \mathrm{even}.
\label{eq:supp_mass_projection}
\end{equation}
The projection uses a Fourier transform along
azimuth followed by contraction with the associated
Legendre basis. Only even degrees through $L=48$ are
retained; conjugate symmetry is used for negative orders.

The projective Gaussian kernel is shape-independent,
so its degree-wise multipliers are precomputed using
$N_q=192$ Gauss--Legendre quadrature nodes:
\begin{equation}
\begin{aligned}
\Lambda[\ell]
&=2\pi\sum_{a=0}^{N_q-1}
\omega_a\kappa_\sigma(\xi_a)P_\ell(\xi_a),\\
\widehat F_{\ell m}
&=\Lambda[\ell]\widehat H_{\ell m},
\end{aligned}
\label{eq:supp_discrete_multipliers}
\end{equation}
where $\sigma=0.12$ radians.

Finally, we synthesize the filtered harmonic expansion
at the vertices of a level-6 icosphere. Antipodal vertices
are identified, leaving $20{,}481$ distinct projective
directions. The field is evaluated directly at these
directions using fast arbitrary-direction spherical-harmonic
synthesis, rather than interpolating values from the
accumulation grid. These field samples form the input
to the candidate readout described in the next section.

\section{Candidate Extraction, Refinement, and Verification}
\label{sec:supp_readout}

We describe how the sampled harmonic symmetry field is
converted into a set of verified reflection planes.

\subsection{Projective Candidate Extraction}
\label{sec:supp_candidate_extraction}

The field is evaluated on a level-6 icosphere with
$40{,}962$ vertices. Identifying antipodal vertices yields
$20{,}481$ distinct directions on $\mathbb{RP}^2$.
We construct projective neighborhoods by mapping the
icosphere edges to their antipodal equivalence classes,
so each direction has five or six neighbors.

Candidate orientations are extracted as local maxima
over this projective adjacency graph. Strict maxima are
retained directly, while equal-valued plateaus are handled
by selecting a deterministic representative of each
eligible isolated maximum.

The extracted peaks are ranked by decreasing field value.
From the strongest $256$ peaks, we discard those below
$1\%$ of the highest peak and retain at most $64$
candidates. Greedy non-maximum suppression (NMS) then
removes candidates within $6^\circ$ of a stronger one,
using the projective angular distance
\begin{equation}
d_{\mathcal M}([\mathbf n_i],[\mathbf n_j])
=\arccos\left(|\mathbf n_i^\top\mathbf n_j|\right).
\label{eq:supp_projective_nms}
\end{equation}
This accounts for the equivalence of opposite normals
and prevents duplicate plane hypotheses.

\subsection{Continuous Angular Refinement}
\label{sec:supp_refinement}

The finite icosphere resolution limits the accuracy of
the extracted orientations. We therefore refine each
surviving candidate directly in the continuous
band-limited harmonic field.

For an initial normal $\mathbf n_0$, we construct an
orthonormal tangent basis $(\mathbf e_1,\mathbf e_2)$.
The field is evaluated at a $3\times3$ stencil of tangent
displacements with $1^\circ$ spacing, mapped onto the
sphere using the exponential map.

Centered finite differences estimate the gradient
$\mathbf g$ and Hessian $\mathbf H$ of the local quadratic
approximation
\begin{equation}
q(\boldsymbol\eta)
=F_L(\mathbf n_0)
+\mathbf g^\top\boldsymbol\eta
+\frac12\boldsymbol\eta^\top
\mathbf H\boldsymbol\eta,
\label{eq:supp_refinement_quadratic}
\end{equation}
where $\boldsymbol\eta\in\mathbb R^2$ denotes a tangent
displacement.

For a well-conditioned, locally concave quadratic,
the proposed update is
$\boldsymbol\eta^*=-\mathbf H^{-1}\mathbf g$.
The displacement is constrained to a $3^\circ$ trust
radius. We evaluate the actual harmonic field at the
full, half, and quarter steps along this displacement,
selecting the trial with the highest field value.

The update is accepted only if it improves the original
field value; otherwise, the initial normal is retained.
A single refinement step is performed per candidate.
Importantly, these evaluations use the continuous harmonic
expansion rather than interpolation of the icosphere
samples.

\subsection{Geometric Verification}
\label{sec:supp_verification}

Maxima of the symmetry field represent strong voting
agreement but need not correspond to genuine reflection
symmetries. We therefore verify each candidate against
the observed point cloud.

Let $\mathbf c$ be the sample centroid. For the original
and refined normals $\mathbf n_0$ and $\mathbf n_1$, the
corresponding planes are
\begin{equation}
\pi_0=(\mathbf n_0,\mathbf n_0^\top\mathbf c),
\qquad
\pi_1=(\mathbf n_1,\mathbf n_1^\top\mathbf c).
\label{eq:supp_candidate_planes}
\end{equation}
Both planes pass through the input centroid.

For each hypothesis, we select $N_v=512$ source points
using deterministic uniformly spaced indices
$i_t=\lfloor tN/N_v\rfloor$, for $t=0,\ldots,N_v-1$.
Each source is reflected through the candidate plane,
and its nearest-neighbor distance to the complete
observed point cloud $\mathcal P$ is computed:
\begin{equation}
e_t(\pi)=\frac{1}{D}
\min_{\mathbf x\in\mathcal P}
\left\|
\mathcal R_\pi(\mathbf p_{i_t})-\mathbf x
\right\|_2,
\label{eq:supp_verification_residual}
\end{equation}
where $D$ is the observed bounding-box diagonal.

The verification score is the $99$th percentile of
these normalized residuals,
\begin{equation}
V(\pi)=Q_{0.99}
\left(\{e_t(\pi)\}_{t=0}^{N_v-1}\right),
\label{eq:supp_verification_score}
\end{equation}
computed using linear quantile interpolation.

We compare $\pi_0$ and $\pi_1$ using the
verification score and select the lower-residual hypothesis. The selected
plane is accepted if
\begin{equation}
\min\{V(\pi_0),V(\pi_1)\}\leq\tau_v,
\qquad \tau_v=0.0147.
\label{eq:supp_verification_acceptance}
\end{equation}
The CPU implementation may certify rejection using distance bounds
without resolving every beyond-threshold residual. Threshold-free
retrieval instead resolves unbounded scores before selecting its plane.

\paragraph{End-to-end procedure.}
The complete FAST detector follows these steps:
\begin{enumerate}
    \item Normalize the input point cloud and sample
    pair-induced reflection hypotheses.
    \item Apply centroid-consistency filtering and
    witness weighting, then construct the angular
    CIC histogram.
    \item Project the histogram onto even spherical
    harmonics, apply the zonal kernel multipliers,
    and synthesize the projective symmetry field.
    \item Extract projective maxima, apply the relative
    peak threshold and candidate cap, and perform
    angular NMS.
    \item Continuously refine each surviving orientation,
    geometrically verify both original and refined
    hypotheses, and return the accepted planes.
\end{enumerate}

\section{Experimental Protocol and Reproducibility}
\label{sec:supp_protocol}

\subsection{Benchmark Construction}
\label{sec:supp_datasets}

\paragraph{PRS.}
We use the annotated PRS-Net benchmark~\cite{Gao2021PRSNet},
with reflection-plane labels from its curated release.
The $850$ labeled shapes are partitioned into disjoint
development (100 shapes) and evaluation (750 shapes) sets.
All $750$ test shapes and their $894$ annotated planes
are retained. Of these shapes, $611$ have one annotated
plane, $134$ have two, and $5$ have three.
The $611$ singly annotated shapes form our single-plane
retrieval benchmark.

\paragraph{Analytic shapes.}
To evaluate objects with larger symmetry groups, we
construct a separate benchmark of procedurally generated
convex polyhedra with analytically known reflection planes.
Table~\ref{tab:supp_analytic_construction} summarizes its
composition.

\begin{table}[t]
\centering
\small
\setlength{\tabcolsep}{5pt}
\begin{tabular}{lrrr}
\toprule
Family & Geometries & Observations & Planes \\
\midrule
Regular prisms & 30 & 30 & 255 \\
Regular pyramids & 30 & 30 & 225 \\
Regular bipyramids & 30 & 30 & 255 \\
Regular frusta & 10 & 10 & 75 \\
Unequal-sided cuboids & 5 & 5 & 15 \\
Platonic solids & 5 & 15 & 162 \\
\midrule
Total & 110 & 120 & 987 \\
\bottomrule
\end{tabular}
\caption{Composition of the analytic benchmark.
The five Platonic solids have three independently
generated observations each; all other geometries
have one observation. Plane counts include all
ground-truth annotations across observations.}
\label{tab:supp_analytic_construction}
\end{table}

Prisms, pyramids, and bipyramids are constructed from
regular $n$-gons with $n=3,\ldots,12$, each at three
fixed height parameters. Frusta use the same polygon
orders with fixed height and unequal top and bottom
radii. We additionally include five unequal-sided
cuboids and all five Platonic solids.

For the polygonal families, the ground-truth planes
comprise the $n$ vertical reflection planes of the
regular polygon. Prisms and bipyramids additionally
have a horizontal reflection plane. Cuboids have three
coordinate-aligned planes, while Platonic-solid planes
are determined from their exact vertex geometry.
The annotations are independently validated by
reflecting polyhedron vertices and checking for a
bijective correspondence.

Each observation is generated with a random rotation,
uniform scale in $[0.7,1.4]$, and translation in
$[-0.75,0.75]^3$. The ground-truth planes are transformed
accordingly. Each Platonic solid is observed three
times with independently generated transformations and
surface samples, yielding $120$ observations of $110$
underlying geometries, with $987$ annotated planes
and between $3$ and $15$ planes per observation.

\subsection{Input Preparation and Noise}
\label{sec:supp_input_preparation}

Each mesh is represented by $N=100{,}000$ surface
samples. Triangles are selected with probability
proportional to their area, followed by uniform
barycentric sampling within each triangle.

Let $D_{\mathrm{clean}}$ denote the axis-aligned
bounding-box diagonal of the clean sampled point
cloud. For noise level $k\in\{1,3,5\}$, each point
is perturbed by
\begin{equation}
\boldsymbol\epsilon\sim
\mathcal N\left(
\mathbf 0,
\frac{(kD_{\mathrm{clean}}/100)^2}{3}\mathbf I_3
\right).
\label{eq:supp_noise}
\end{equation}
Thus, the expected squared Euclidean displacement
corresponds to an RMS displacement of
$k\%$ of $D_{\mathrm{clean}}$.
The same clean samples and Gaussian draws are reused
across noise levels, with only their magnitude scaled.

Each noisy observation is subsequently centered by
its own sample mean and normalized by its observed
bounding-box diagonal. Consequently, noise magnitude
is defined relative to the clean shape, whereas
detector normalization and geometric matching use
the observed shape extent.

All point-cloud methods receive corresponding
observations, with method-specific preprocessing
applied as required. Mesh-based methods retain their
native input representation. In particular, Kazhdan
uses subdivided meshes with perturbed vertices under
noise, rather than independently perturbed surface
samples.

\subsection{Evaluation Metrics}
\label{sec:supp_metrics}

\paragraph{Multi-plane detection.}
We match predicted planes to ground-truth annotations
using one-to-one Hungarian assignment~\cite{Kuhn1955Hungarian}.
For two planes $(\mathbf n,\ell)$ and
$(\mathbf m,b)$ with unit normals, we define
\begin{equation}
\begin{aligned}
e_{\mathrm{ang}}
&=\arccos\left(|\mathbf n^\top\mathbf m|\right),\\
e_{\mathrm{off}}
&=\frac{|s\ell-b|}{D},
\qquad
s=\operatorname{sign}(\mathbf n^\top\mathbf m),
\end{aligned}
\label{eq:supp_matching_errors}
\end{equation}
where $D$ is the observed bounding-box diagonal.
The sign adjustment accounts for the equivalence
$(\mathbf n,\ell)\sim(-\mathbf n,-\ell)$.

A match is feasible only when
$e_{\mathrm{ang}}\leq5^\circ$ and
$e_{\mathrm{off}}\leq0.02$.
The assignment maximizes feasible match cardinality
and then minimizes the sum of normalized geometric
errors. Unmatched predictions and annotations count
as false positives and false negatives, respectively.

For each dataset and noise level, counts are pooled
over all observations before computing
\begin{equation}
\begin{aligned}
P=\frac{TP}{TP+FP}, \\
R=\frac{TP}{TP+FN},\\
F_1=\frac{2TP}{2TP+FP+FN}.
\end{aligned}
\end{equation}
Localization errors are reported as medians over
matched true-positive planes.

\paragraph{Single-plane retrieval.}
On the $611$ singly annotated PRS shapes, each method
selects its available candidate with the lowest
geometric $99$th-percentile verification residual.
This common ranking is applied without an acceptance
threshold, returning at most one plane per shape.
Missing candidates count as unsuccessful retrievals.
A retrieval succeeds if its selected plane satisfies
the same $5^\circ$ angular and $0.02D$ offset tolerances.
Success rates are averaged over all $611$ shapes.

\subsection{Hyperparameters and Selection}
\label{sec:supp_hyperparameters}

Table~\ref{tab:supp_hyperparameters} summarizes the
detector configuration used throughout the experiments.
The centroid gate and geometric acceptance threshold
were calibrated on the clean validation set by optimizing pooled
detection $F_1$. The bandlimit $L=48$ was selected using the
accuracy--efficiency study reported in
Section~\ref{sec:supp_resolution}. The reported FAST configuration uses
the same parameters on both benchmarks and at every
noise level, without dataset-specific or noise-specific
threshold adaptation.

\begin{table}[t]
\centering
\small
\setlength{\tabcolsep}{5pt}
\begin{tabular}{ll}
\toprule
Parameter & Value \\
\midrule
Surface samples & $100{,}000$ \\
Sampled point pairs & $250{,}000$ \\
Witnesses per pair & $2$ \\
Centroid gate $\delta$ & $0.0108$ \\
Witness tolerance $\tau_w$ & $0.01$ \\
Occupancy resolution & $96^3$ \\
\midrule
Gaussian bandwidth $\sigma$ & $0.12$ rad \\
Harmonic bandlimit $L$ & $48$ \\
Angular histogram & $50\times193$ \\
Kernel quadrature nodes & $192$ \\
Projective directions & $20{,}481$ \\
\midrule
Peak threshold & $1\%$ of maximum \\
Initial peak limit & $256$ \\
Candidate limit & $64$ \\
Projective NMS radius & $6^\circ$ \\
Refinement stencil & $3\times3$, $1^\circ$ spacing \\
Refinement trust radius & $3^\circ$ \\
Refinement steps & $1$ \\
Verification sources & $512$ \\
Verification statistic & $99$th percentile \\
Verification cutoff $\tau_v$ & $0.0147$ \\
\bottomrule
\end{tabular}
\caption{Default FAST configuration; ablations change the specified
settings. Distance thresholds are expressed in
normalized coordinates.}
\label{tab:supp_hyperparameters}
\end{table}

\subsection{Baseline Configurations}
\label{sec:supp_baselines}

We evaluate geometric, spectral, and learning-based
detectors using their released implementations or
documented adaptations. Pretrained methods use their
released checkpoints without additional training.
Each baseline retains its native input representation,
preprocessing, and candidate-generation procedure;
all predicted planes are transformed into the common
observation coordinate system before evaluation.

For multi-plane detection, methods retain their native
candidate scores and acceptance mechanisms. Where a
scalar acceptance threshold is applicable, it is
calibrated on the same clean validation dataset and held fixed across
benchmarks and noise levels.

Two variants require clarification. Our PRST baseline
uses a corrected third-party Python implementation
of Podolak et al.~\cite{Podolak:2006:APS}, evaluated
using its own calibrated confidence criterion.
The additional ``Langevin + verifier'' experiment
applies our geometric verification procedure to the
proposals of Je et al.~\cite{Je2024Langevin}, with a
separately calibrated acceptance threshold. It is
reported separately from native Langevin.

Methods restricted to one or a small fixed number
of planes are additionally assessed in the
single-plane experiment. There, candidate selection
uses the common geometric ranking described in
Section~\ref{sec:supp_metrics}, rather than each
method's native multi-plane acceptance threshold.
Candidate-generation procedures themselves remain
method-specific, including upstream filtering and candidate budgets.
The common ranking preserves each candidate's predicted offset.

\subsection{Runtime and Operator Evaluation}
\label{sec:supp_timing}

\paragraph{Detector runtime.}
FAST is evaluated on an NVIDIA RTX A5000 GPU and
an Intel Core i7-13700K CPU. For each observation, we record a single timed run after warmup and report the median across the complete cohort.

FAST's main-paper runtime is the outer detector-call wall time
from a prepared point cloud to the returned detector output.
It includes input normalization, pair voting, harmonic processing,
candidate extraction, refinement, geometric verification, and
returned diagnostic construction. Input loading, surface sampling,
reusable initialization, compilation, warmup, ground-truth matching,
and serialization are excluded.

CPU methods use multithreading where supported. FAST uses its
frozen eight-logical-CPU implementation; the final Kazhdan and
SymSeg measurements use eight-thread configurations. Korman PAS
and Hruda retain their native parallel implementations, while the
PRST port and PCA retain serial numerical paths.

\paragraph{Single-plane retrieval runtime.}
Runtimes use all 611 clean singly annotated PRS shapes.
FAST is timed as a complete detector-plus-selection call.
For each baseline and shape, the recorded native detector
time is added to the independently measured common-p99
verification and candidate-selection time; the reported value is
the median of these per-shape totals.

\paragraph{Controlled harmonic-operator comparison.}
We compare harmonic field construction and evaluation against
direct summation of the original projective geodesic Gaussian,
using identical normalized deposited histograms and projective
query directions. The fixed clean cohort contains ten PRS and ten
\textsc{Analytic} observations. We cross
$L\in\{16,32,48,64,80,96\}$ with icosphere levels 4, 5, and 6,
giving $N_u\in\{1281,5121,20481\}$ queries.
The histogram has $(L+2)(4L+1)$ bins and kernel quadrature uses
$\max(32,4L)$ nodes.

Both methods use native vectorized array operations on one CPU
thread with affinity restricted to the same logical CPU.
The harmonic method uses a longitude FFT, latitude contraction,
degree-wise filtering, and fast arbitrary-direction synthesis;
DIRECT evaluates
\begin{equation}
F_{\mathrm{direct}}(\mathbf u)
=\sum_{q,p}H[q,p]\,
\exp\!\left[-\frac{\arccos^2(|\mathbf u^\top\mathbf v_{qp}|)}
{\sigma^2}\right]
\label{eq:supp_direct_gaussian}
\end{equation}
using vectorized dot products, inverse cosines, exponentials,
and reductions in bounded-memory query tiles.

Each timed call includes per-observation representation construction
and evaluation at all query directions. Voting, deposition, peak extraction,
refinement, verification, reusable precomputation, and validation
are excluded. After five initial warm calls and one warm call per
observation, we record seven repetitions per observation. We first
take per-observation medians, then the median over the 20 observations.
Combined time is timed directly, rather than obtained by adding
separately aggregated construction and query medians.

Table~\ref{tab:supp_operator_current} reports level-6 results.
Relative field error pools all observation--query values:
$\|F_{\mathrm{harmonic}}-F_{\mathrm{direct}}\|_2/
\|F_{\mathrm{direct}}\|_2$.
The harmonic field is band-limited, whereas DIRECT uses the
untruncated Gaussian; low-$L$ discrepancies therefore include
truncation error. At $L=48$, the measured combined-time speedup
is approximately $1{,}405\times$, with relative field error
$1.56\times10^{-5}$. Fits across all 18 settings use separate
nonnegative models $a+bL^3+cN_u$ and $a'+b'N_uL^2$.

\begin{table}[t]
\centering
\small
\setlength{\tabcolsep}{3pt}
\begin{tabular}{rrrrr}
\toprule
$L$ & Harmonic (ms) & DIRECT (ms) & Speedup & Relative error \\
\midrule
16 & 0.531 & 189.492 & 356.8$\times$ & $9.53\!\times\!10^{-2}$ \\
32 & 0.592 & 618.537 & 1044.0$\times$ & $2.44\!\times\!10^{-3}$ \\
48 & 0.830 & 1166.873 & 1405.3$\times$ & $1.56\!\times\!10^{-5}$ \\
64 & 0.973 & 1983.986 & 2038.1$\times$ & $2.58\!\times\!10^{-6}$ \\
80 & 1.278 & 3008.887 & 2354.3$\times$ & $2.54\!\times\!10^{-6}$ \\
96 & 1.719 & 4231.817 & 2462.1$\times$ & $2.48\!\times\!10^{-6}$ \\
\bottomrule
\end{tabular}
\caption{Single-threaded CPU field construction and evaluation at
20,481 projective directions, pooled over 20 clean observations.
Speedups use the unrounded combined-time medians. Relative errors
compare the harmonic approximation with the original Gaussian.}
\label{tab:supp_operator_current}
\end{table}

\begingroup
\setlength{\dbltextfloatsep}{10pt plus 3pt minus 3pt}
\setlength{\textfloatsep}{9pt plus 2pt minus 2pt}
\renewcommand{\dbltopfraction}{0.92}
\renewcommand{\topfraction}{0.92}
\renewcommand{\textfraction}{0.06}
\renewcommand{\floatpagefraction}{0.83}
\renewcommand{\dblfloatpagefraction}{0.83}
\setcounter{dbltopnumber}{2}
\setcounter{topnumber}{2}
\section{Complete Quantitative Results}
\label{sec:supp_complete_results}

We report precision and recall in addition to the $F_1$ scores
presented in the main paper. All multi-plane metrics are computed
from pooled true-positive, false-positive, and false-negative counts
under the assignment protocol of Section~\ref{sec:supp_metrics}.
Unless stated otherwise, FAST's accuracy and localization tables use
GPU predictions. 

\subsection{Precision, Recall, and $F_1$}
\label{sec:supp_full_prf}

Tables~\ref{tab:supp_prs_prf} and~\ref{tab:supp_analytic_prf}
report the complete multi-plane results for every method and noise
level. FAST achieves clean $(P,R)=(0.839,0.885)$ on PRS and
$(0.941,0.952)$ on \textsc{Analytic}. On PRS, recall remains above
$0.90$ at $5\%$ noise, while precision decreases because of
additional false-positive planes. On \textsc{Analytic}, both
precision and recall decline with stronger perturbations.

\begin{table*}[!t]
\centering
{\fontsize{7.9pt}{9.2pt}\selectfont\setlength{\tabcolsep}{2.5pt}\renewcommand{\arraystretch}{1.04}
\begin{tabular}{lcccccccccccc}
\toprule
Method & \multicolumn{3}{c}{Clean} & \multicolumn{3}{c}{1\%} & \multicolumn{3}{c}{3\%} & \multicolumn{3}{c}{5\%} \\
 & P & R & $F_1$ & P & R & $F_1$ & P & R & $F_1$ & P & R & $F_1$ \\
\midrule
FAST (ours) & 0.839 & 0.885 & 0.861 & 0.776 & 0.938 & 0.850 & 0.621 & 0.947 & 0.751 & 0.504 & 0.906 & 0.647 \\
Korman PAS & 0.770 & 0.833 & 0.800 & 0.785 & 0.737 & 0.760 & 0.734 & 0.102 & 0.179 & 0.500 & 0.001 & 0.002 \\
EBP & 0.649 & 0.941 & 0.768 & 0.575 & 0.719 & 0.639 & 0.416 & 0.415 & 0.416 & 0.289 & 0.199 & 0.236 \\
Kazhdan & 0.652 & 0.915 & 0.761 & 0.738 & 0.956 & 0.833 & 0.611 & 0.962 & 0.747 & 0.483 & 0.966 & 0.644 \\
E3Sym & 0.604 & 0.815 & 0.694 & 0.602 & 0.718 & 0.655 & 0.564 & 0.494 & 0.527 & 0.495 & 0.304 & 0.377 \\
SymSeg & 0.562 & 0.739 & 0.639 & 0.682 & 0.017 & 0.033 & -- & 0.000 & 0.000 & -- & 0.000 & 0.000 \\
PRST & 0.602 & 0.742 & 0.665 & 0.498 & 0.739 & 0.595 & 0.296 & 0.672 & 0.411 & 0.170 & 0.582 & 0.263 \\
Langevin + verifier & 0.577 & 0.752 & 0.653 & 0.489 & 0.813 & 0.611 & 0.386 & 0.789 & 0.519 & 0.318 & 0.736 & 0.444 \\
Langevin & 0.241 & 0.897 & 0.380 & 0.246 & 0.869 & 0.384 & 0.242 & 0.802 & 0.372 & 0.229 & 0.737 & 0.349 \\
\bottomrule
\end{tabular}
\par}
\caption{Multi-plane precision (P), recall (R), and micro-$F_1$ on PRS test set. Values are computed from pooled TP/FP/FN counts. A dash denotes undefined precision when no planes are predicted.}\label{tab:supp_prs_prf}
\vspace{8pt}
{\fontsize{7.9pt}{9.2pt}\selectfont\setlength{\tabcolsep}{2.5pt}\renewcommand{\arraystretch}{1.04}
\begin{tabular}{lcccccccccccc}
\toprule
Method & \multicolumn{3}{c}{Clean} & \multicolumn{3}{c}{1\%} & \multicolumn{3}{c}{3\%} & \multicolumn{3}{c}{5\%} \\
 & P & R & $F_1$ & P & R & $F_1$ & P & R & $F_1$ & P & R & $F_1$ \\
\midrule
FAST (ours) & 0.941 & 0.952 & 0.947 & 0.868 & 0.881 & 0.875 & 0.785 & 0.744 & 0.764 & 0.679 & 0.609 & 0.642 \\
Korman PAS & 0.998 & 0.819 & 0.899 & 0.997 & 0.798 & 0.887 & 1.000 & 0.016 & 0.032 & -- & 0.000 & 0.000 \\
EBP & 0.957 & 0.898 & 0.926 & 0.825 & 0.287 & 0.426 & 1.000 & 0.008 & 0.016 & -- & 0.000 & 0.000 \\
Kazhdan & 0.815 & 0.963 & 0.882 & 0.796 & 0.817 & 0.806 & 0.721 & 0.811 & 0.763 & 0.679 & 0.806 & 0.737 \\
E3Sym & 0.874 & 0.161 & 0.272 & 0.768 & 0.110 & 0.193 & 0.482 & 0.041 & 0.075 & 0.418 & 0.023 & 0.044 \\
SymSeg & 0.993 & 0.721 & 0.836 & 0.980 & 0.485 & 0.649 & -- & 0.000 & 0.000 & -- & 0.000 & 0.000 \\
PRST & 0.832 & 0.260 & 0.397 & 0.714 & 0.281 & 0.403 & 0.452 & 0.229 & 0.304 & 0.243 & 0.181 & 0.208 \\
Langevin + verifier & 0.782 & 0.393 & 0.523 & 0.684 & 0.328 & 0.444 & 0.494 & 0.233 & 0.317 & 0.313 & 0.137 & 0.190 \\
Langevin & 0.550 & 0.407 & 0.468 & 0.482 & 0.334 & 0.395 & 0.355 & 0.233 & 0.282 & 0.224 & 0.137 & 0.170 \\
\bottomrule
\end{tabular}
\par}
\caption{Multi-plane precision (P), recall (R), and micro-$F_1$ on \textsc{Analytic}. Values are computed from pooled TP/FP/FN counts. A dash denotes undefined precision when no planes are predicted.}\label{tab:supp_analytic_prf}

\end{table*}

\subsection{Localization and Single-Plane Retrieval}
\label{sec:supp_full_localization}

Table~\ref{tab:supp_localization} provides angular and normalized
offset errors at each noise level, computed only over matched true
positives. These conditional localization errors should be
interpreted together with detection precision and recall: a method
returning very few planes can exhibit small localization error on
the few predictions that remain matched.

Single-plane retrieval is a separate top-1 task rather than
multi-plane detection. Table~\ref{tab:supp_retrieval} reproduces the
retrieval success rates using the common lowest-p99 selection rule;
we do not assign multi-plane precision or recall to this setting.

\begin{table*}[!t]
\centering
{\fontsize{8.2pt}{9.5pt}\selectfont\setlength{\tabcolsep}{3pt}\renewcommand{\arraystretch}{1.01}
\begin{tabular}{lcccc}
\toprule
Method & Clean & 1\% & 3\% & 5\% \\
\midrule
\multicolumn{5}{l}{\emph{PRS}} \\
\midrule
FAST (ours) & 0.191 / 0.00093 & 0.303 / 0.00094 & 0.506 / 0.00104 & 0.699 / 0.00108 \\
Korman PAS & 0.513 / 0.00185 & 0.484 / 0.00173 & 0.525 / 0.00139 & 3.254 / 0.00255 \\
EBP & 0.294 / 0.00152 & 0.786 / 0.00226 & 0.919 / 0.00246 & 1.005 / 0.00263 \\
Kazhdan & 1.321 / 0.00137 & 1.381 / 0.00085 & 1.390 / 0.00057 & 1.419 / 0.00050 \\
E3Sym & 0.782 / 0.00377 & 1.072 / 0.00402 & 1.519 / 0.00493 & 1.959 / 0.00483 \\
SymSeg & 0.027 / 0.00066 & 0.186 / 0.00023 & -- & -- \\
PRST & 1.643 / 0.00408 & 1.708 / 0.00418 & 2.103 / 0.00424 & 2.515 / 0.00448 \\
Langevin + verifier & 0.274 / 0.00737 & 0.337 / 0.00725 & 0.506 / 0.00673 & 0.729 / 0.00641 \\
Langevin & 0.294 / 0.00723 & 0.367 / 0.00697 & 0.520 / 0.00668 & 0.731 / 0.00640 \\
\midrule
\multicolumn{5}{l}{\emph{\textsc{Analytic}}} \\
\midrule
FAST (ours) & 0.329 / 0.00073 & 0.496 / 0.00096 & 1.121 / 0.00171 & 1.859 / 0.00227 \\
Korman PAS & 0.583 / 0.00099 & 0.671 / 0.00112 & 0.995 / 0.00181 & -- \\
EBP & 0.452 / 0.00108 & 1.277 / 0.00258 & 1.408 / 0.00292 & -- \\
Kazhdan & 1.466 / 0.00225 & 1.517 / 0.00302 & 1.538 / 0.00281 & 1.651 / 0.00265 \\
E3Sym & 1.217 / 0.00282 & 1.714 / 0.00390 & 2.669 / 0.00610 & 3.347 / 0.00510 \\
SymSeg & 0.004 / 0.00001 & 0.363 / 0.00059 & -- & -- \\
PRST & 1.634 / 0.00395 & 1.803 / 0.00418 & 2.434 / 0.00465 & 3.020 / 0.00527 \\
Langevin + verifier & 0.832 / 0.00371 & 1.121 / 0.00424 & 1.932 / 0.00659 & 2.175 / 0.00647 \\
Langevin & 0.860 / 0.00377 & 1.128 / 0.00424 & 1.932 / 0.00659 & 2.175 / 0.00647 \\
\bottomrule
\end{tabular}
\par}
\caption{Median matched-true-positive localization errors, angular degrees / normalized offset $e_{\mathrm{off}}$. Unmatched predictions are excluded; dashes indicate no matches. FAST GPU localization is shown.}\label{tab:supp_localization}

\end{table*}

\begin{table}[t]
\centering
{\fontsize{8.6pt}{10.0pt}\selectfont\setlength{\tabcolsep}{4pt}\renewcommand{\arraystretch}{1.03}
\begin{tabular}{lcccc}
\toprule
Method & Clean & 1\% & 3\% & 5\% \\
\midrule
FAST (ours) & 0.949 & 0.917 & 0.848 & 0.753 \\
EBP & 0.915 & 0.876 & 0.805 & 0.671 \\
Kazhdan & 0.903 & 0.943 & 0.918 & 0.879 \\
Korman PAS & 0.894 & 0.910 & 0.813 & 0.689 \\
PCA & 0.900 & 0.885 & 0.836 & 0.768 \\
SymSeg & 0.843 & 0.705 & 0.489 & 0.473 \\
Langevin & 0.810 & 0.769 & 0.714 & 0.651 \\
PRST & 0.782 & 0.748 & 0.692 & 0.604 \\
E3Sym & 0.736 & 0.664 & 0.475 & 0.296 \\
PRS-Net & 0.406 & 0.401 & 0.390 & 0.385 \\
Hruda & 0.789 & 0.748 & 0.221 & 0.047 \\
\bottomrule
\end{tabular}
\par}
\caption{Top-1 retrieval success on the 611 singly annotated PRS shapes. Each method returns its lowest-p99 available candidate, without a detection threshold.}\label{tab:supp_retrieval}
\end{table}

\section{Statistical Reliability and Confidence Intervals}
\label{sec:supp_statistics}

\subsection{Paired Bootstrap Protocol}
\label{sec:supp_bootstrap_protocol}

We quantify FAST-minus-baseline differences using $200{,}000$
paired bootstrap resamples. The resampling units are $750$ PRS
shapes, $110$ underlying \textsc{Analytic} geometries (keeping their
$120$ observations grouped), and $611$ shapes for single-plane
retrieval. For each resample, we recompute pooled micro-$F_1$ or
top-1 retrieval success, respectively. Identical bootstrap weights
are applied to FAST and its comparator. The reported comparison
families use GPU FAST outcomes and the corresponding saved
baseline outcomes.

We report ordinary percentile $95\%$ confidence intervals and
Bonferroni-adjusted intervals for the pre-specified families of
$64$ multi-plane comparisons and $40$ retrieval comparisons.
For a family of size $m$, the adjusted bounds are the empirical
quantiles at $0.05/(2m)$ and $1-0.05/(2m)$. Differences and
intervals are expressed as fractions, not percentage points.
An interval containing zero does not establish an advantage for
either method.

\subsection{Complete Multi-Plane Comparisons}
\label{sec:supp_multi_ci}

Tables~\ref{tab:supp_prs_ci} and~\ref{tab:supp_analytic_ci}
contain all $64$ comparisons. After familywise correction, FAST
has a supported advantage in $57$ comparisons, six remain
unresolved, and Kazhdan is favored in one comparison
(\textsc{Analytic}, $5\%$ noise). On clean PRS, FAST's advantage
over Korman PAS is supported; on clean \textsc{Analytic}, its
smaller advantage over EBP is not statistically resolved.

\begin{table*}[!t]
\centering
\begin{minipage}[t]{0.486\textwidth}\centering\vspace{0pt}
{\fontsize{7.1pt}{8.6pt}\selectfont\setlength{\tabcolsep}{1.3pt}\renewcommand{\arraystretch}{0.98}
\resizebox{\linewidth}{!}{\begin{tabular}{clrrr}
\toprule
Noise & Comparator & $\Delta$ & 95\% CI & Adjusted 95\% CI \\
\midrule
0\% & EBP & +0.0932 & [+0.0694, +0.1171] & [+0.0529, +0.1347] \\
0\% & E3Sym & +0.1669 & [+0.1432, +0.1909] & [+0.1267, +0.2073] \\
0\% & Kazhdan & +0.0999 & [+0.0745, +0.1264] & [+0.0587, +0.1464] \\
0\% & Korman PAS & +0.0610 & [+0.0419, +0.0804] & [+0.0284, +0.0940] \\
0\% & Langevin & +0.4816 & [+0.4618, +0.5014] & [+0.4472, +0.5146] \\
0\% & Langevin + verifier & +0.2081 & [+0.1887, +0.2276] & [+0.1749, +0.2414] \\
0\% & PRST & +0.1965 & [+0.1684, +0.2247] & [+0.1478, +0.2443] \\
0\% & SymSeg & +0.2225 & [+0.1963, +0.2489] & [+0.1781, +0.2674] \\
\midrule
1\% & EBP & +0.2105 & [+0.1839, +0.2369] & [+0.1656, +0.2553] \\
1\% & E3Sym & +0.1945 & [+0.1696, +0.2199] & [+0.1520, +0.2384] \\
1\% & Kazhdan & +0.0163 & [-0.0087, +0.0424] & [-0.0246, +0.0619] \\
1\% & Korman PAS & +0.0895 & [+0.0672, +0.1122] & [+0.0517, +0.1288] \\
1\% & Langevin & +0.4658 & [+0.4469, +0.4848] & [+0.4334, +0.4978] \\
1\% & Langevin + verifier & +0.2390 & [+0.2219, +0.2560] & [+0.2099, +0.2681] \\
1\% & PRST & +0.2544 & [+0.2279, +0.2811] & [+0.2090, +0.3005] \\
1\% & SymSeg & +0.8169 & [+0.7918, +0.8410] & [+0.7734, +0.8570] \\
\midrule
3\% & EBP & +0.3349 & [+0.3068, +0.3635] & [+0.2874, +0.3844] \\
3\% & E3Sym & +0.2237 & [+0.1924, +0.2557] & [+0.1709, +0.2777] \\
3\% & Kazhdan & +0.0034 & [-0.0221, +0.0291] & [-0.0398, +0.0488] \\
3\% & Korman PAS & +0.5718 & [+0.5279, +0.6156] & [+0.4963, +0.6458] \\
3\% & Langevin & +0.3789 & [+0.3545, +0.4035] & [+0.3372, +0.4206] \\
3\% & Langevin + verifier & +0.2320 & [+0.2102, +0.2538] & [+0.1945, +0.2685] \\
3\% & PRST & +0.3393 & [+0.3127, +0.3659] & [+0.2940, +0.3842] \\
3\% & SymSeg & +0.7506 & [+0.7252, +0.7757] & [+0.7060, +0.7927] \\
\midrule
5\% & EBP & +0.4116 & [+0.3765, +0.4477] & [+0.3507, +0.4740] \\
5\% & E3Sym & +0.2705 & [+0.2346, +0.3069] & [+0.2090, +0.3326] \\
5\% & Kazhdan & +0.0032 & [-0.0212, +0.0274] & [-0.0377, +0.0440] \\
5\% & Korman PAS & +0.6452 & [+0.6165, +0.6740] & [+0.5952, +0.6933] \\
5\% & Langevin & +0.2983 & [+0.2726, +0.3243] & [+0.2549, +0.3431] \\
5\% & Langevin + verifier & +0.2036 & [+0.1803, +0.2274] & [+0.1637, +0.2438] \\
5\% & PRST & +0.3843 & [+0.3576, +0.4112] & [+0.3383, +0.4297] \\
5\% & SymSeg & +0.6475 & [+0.6198, +0.6755] & [+0.5994, +0.6941] \\
\bottomrule
\end{tabular}}\par}
\caption{All 32 paired multi-plane comparisons on PRS, $\Delta$ = FAST minus comparator micro-$F_1$. Familywise-adjusted intervals use all 64 multi-plane contrasts.}\label{tab:supp_prs_ci}
\end{minipage}
\hfill
\begin{minipage}[t]{0.486\textwidth}\centering\vspace{0pt}
{\fontsize{7.1pt}{8.6pt}\selectfont\setlength{\tabcolsep}{1.3pt}\renewcommand{\arraystretch}{0.98}
\resizebox{\linewidth}{!}{\begin{tabular}{clrrr}
\toprule
Noise & Comparator & $\Delta$ & 95\% CI & Adjusted 95\% CI \\
\midrule
0\% & EBP & +0.0203 & [-0.0105, +0.0497] & [-0.0324, +0.0698] \\
0\% & E3Sym & +0.6746 & [+0.6190, +0.7273] & [+0.5816, +0.7639] \\
0\% & Kazhdan & +0.0641 & [+0.0396, +0.0899] & [+0.0232, +0.1091] \\
0\% & Korman PAS & +0.0473 & [+0.0267, +0.0715] & [+0.0137, +0.0919] \\
0\% & Langevin & +0.4786 & [+0.4225, +0.5333] & [+0.3832, +0.5682] \\
0\% & Langevin + verifier & +0.4234 & [+0.3646, +0.4812] & [+0.3241, +0.5187] \\
0\% & PRST & +0.5500 & [+0.5021, +0.5964] & [+0.4673, +0.6274] \\
0\% & SymSeg & +0.1109 & [+0.0873, +0.1365] & [+0.0707, +0.1542] \\
\midrule
1\% & EBP & +0.4492 & [+0.3539, +0.5464] & [+0.2895, +0.6150] \\
1\% & E3Sym & +0.6817 & [+0.6247, +0.7362] & [+0.5850, +0.7727] \\
1\% & Kazhdan & +0.0684 & [+0.0334, +0.1042] & [+0.0078, +0.1302] \\
1\% & Korman PAS & -0.0121 & [-0.0405, +0.0168] & [-0.0622, +0.0386] \\
1\% & Langevin & +0.4798 & [+0.4140, +0.5450] & [+0.3680, +0.5884] \\
1\% & Langevin + verifier & +0.4313 & [+0.3632, +0.4992] & [+0.3162, +0.5449] \\
1\% & PRST & +0.4719 & [+0.4255, +0.5172] & [+0.3913, +0.5495] \\
1\% & SymSeg & +0.2258 & [+0.1816, +0.2774] & [+0.1522, +0.3168] \\
\midrule
3\% & EBP & +0.7477 & [+0.6917, +0.7995] & [+0.6527, +0.8336] \\
3\% & E3Sym & +0.6890 & [+0.6306, +0.7436] & [+0.5903, +0.7799] \\
3\% & Kazhdan & +0.0008 & [-0.0346, +0.0338] & [-0.0601, +0.0556] \\
3\% & Korman PAS & +0.7319 & [+0.6719, +0.7873] & [+0.6294, +0.8228] \\
3\% & Langevin & +0.4823 & [+0.4219, +0.5422] & [+0.3833, +0.5833] \\
3\% & Langevin + verifier & +0.4472 & [+0.3836, +0.5114] & [+0.3449, +0.5558] \\
3\% & PRST & +0.4598 & [+0.4036, +0.5150] & [+0.3652, +0.5531] \\
3\% & SymSeg & +0.7638 & [+0.7123, +0.8118] & [+0.6759, +0.8449] \\
\midrule
5\% & EBP & +0.6421 & [+0.5922, +0.6966] & [+0.5603, +0.7329] \\
5\% & E3Sym & +0.5979 & [+0.5473, +0.6536] & [+0.5147, +0.6927] \\
5\% & Kazhdan & -0.0949 & [-0.1539, -0.0350] & [-0.1934, -0.0003] \\
5\% & Korman PAS & +0.6421 & [+0.5922, +0.6966] & [+0.5603, +0.7329] \\
5\% & Langevin & +0.4724 & [+0.4294, +0.5207] & [+0.4016, +0.5564] \\
5\% & Langevin + verifier & +0.4517 & [+0.4099, +0.4983] & [+0.3827, +0.5326] \\
5\% & PRST & +0.4346 & [+0.3956, +0.4761] & [+0.3688, +0.5061] \\
5\% & SymSeg & +0.6421 & [+0.5922, +0.6966] & [+0.5603, +0.7329] \\
\bottomrule
\end{tabular}}\par}
\caption{All 32 paired multi-plane comparisons on \textsc{Analytic}, $\Delta$ = FAST minus comparator micro-$F_1$. Familywise-adjusted intervals use all 64 multi-plane contrasts.}\label{tab:supp_analytic_ci}
\end{minipage}
\end{table*}

\subsection{Complete Single-Plane Comparisons}
\label{sec:supp_single_ci}

Table~\ref{tab:supp_retrieval_ci} reports all $40$ retrieval
comparisons. FAST has $28$ supported advantages, ten unresolved
comparisons, and two supported disadvantages, both against
Kazhdan at $3\%$ and $5\%$ noise.

\subsection{Interpretation}
\label{sec:supp_stat_interpretation}

The results support FAST's broad performance advantage while also
identifying close and noise-dependent comparisons that the point
estimates alone cannot resolve.

\begin{table*}[!t]
\centering
\begin{minipage}[t]{0.486\textwidth}\centering\vspace{0pt}
{\fontsize{7.2pt}{8.6pt}\selectfont\setlength{\tabcolsep}{1.2pt}\renewcommand{\arraystretch}{0.99}
\resizebox{\linewidth}{!}{\begin{tabular}{@{}clrrr@{}}
\toprule
Noise & Comparator & $\Delta$ & 95\% CI & Adjusted 95\% CI \\
\midrule
0\% & EBP & +0.0344 & [+0.0131, +0.0573] & [-0.0016, +0.0720] \\
0\% & E3Sym & +0.2128 & [+0.1768, +0.2504] & [+0.1538, +0.2733] \\
0\% & Kazhdan & +0.0458 & [+0.0245, +0.0671] & [+0.0115, +0.0835] \\
0\% & Korman PAS & +0.0556 & [+0.0311, +0.0802] & [+0.0164, +0.0966] \\
0\% & Langevin & +0.1391 & [+0.1080, +0.1718] & [+0.0884, +0.1931] \\
0\% & PCA & +0.0491 & [+0.0262, +0.0736] & [+0.0098, +0.0900] \\
0\% & PRS-Net & +0.5434 & [+0.5008, +0.5859] & [+0.4730, +0.6121] \\
0\% & PRST & +0.1669 & [+0.1342, +0.2013] & [+0.1129, +0.2226] \\
0\% & SymSeg & +0.1064 & [+0.0769, +0.1375] & [+0.0589, +0.1571] \\
0\% & Hruda & +0.1604 & [+0.1277, +0.1931] & [+0.1064, +0.2144] \\
\midrule
1\% & EBP & +0.0409 & [+0.0131, +0.0687] & [-0.0049, +0.0884] \\
1\% & E3Sym & +0.2520 & [+0.2128, +0.2913] & [+0.1866, +0.3175] \\
1\% & Kazhdan & -0.0262 & [-0.0524, +0.0000] & [-0.0687, +0.0180] \\
1\% & Korman PAS & +0.0065 & [-0.0196, +0.0327] & [-0.0360, +0.0507] \\
1\% & Langevin & +0.1473 & [+0.1113, +0.1833] & [+0.0884, +0.2062] \\
1\% & PCA & +0.0311 & [+0.0049, +0.0573] & [-0.0115, +0.0736] \\
1\% & PRS-Net & +0.5155 & [+0.4714, +0.5597] & [+0.4419, +0.5876] \\
1\% & PRST & +0.1686 & [+0.1309, +0.2062] & [+0.1064, +0.2291] \\
1\% & SymSeg & +0.2111 & [+0.1735, +0.2488] & [+0.1506, +0.2733] \\
1\% & Hruda & +0.1686 & [+0.1326, +0.2046] & [+0.1097, +0.2291] \\
\bottomrule
\end{tabular}}\par}
\end{minipage}
\hfill
\begin{minipage}[t]{0.486\textwidth}\centering\vspace{0pt}
{\fontsize{7.2pt}{8.6pt}\selectfont\setlength{\tabcolsep}{1.2pt}\renewcommand{\arraystretch}{0.99}
\resizebox{\linewidth}{!}{\begin{tabular}{@{}clrrr@{}}
\toprule
Noise & Comparator & $\Delta$ & 95\% CI & Adjusted 95\% CI \\
\midrule
3\% & EBP & +0.0426 & [+0.0065, +0.0786] & [-0.0164, +0.1015] \\
3\% & E3Sym & +0.3732 & [+0.3290, +0.4173] & [+0.2995, +0.4468] \\
3\% & Kazhdan & -0.0704 & [-0.1031, -0.0376] & [-0.1260, -0.0164] \\
3\% & Korman PAS & +0.0344 & [+0.0016, +0.0671] & [-0.0180, +0.0884] \\
3\% & Langevin & +0.1342 & [+0.0933, +0.1751] & [+0.0671, +0.2029] \\
3\% & PCA & +0.0115 & [-0.0196, +0.0426] & [-0.0409, +0.0638] \\
3\% & PRS-Net & +0.4583 & [+0.4124, +0.5041] & [+0.3830, +0.5336] \\
3\% & PRST & +0.1555 & [+0.1162, +0.1948] & [+0.0917, +0.2209] \\
3\% & SymSeg & +0.3584 & [+0.3126, +0.4043] & [+0.2848, +0.4321] \\
3\% & Hruda & +0.6268 & [+0.5859, +0.6661] & [+0.5597, +0.6923] \\
\midrule
5\% & EBP & +0.0818 & [+0.0442, +0.1195] & [+0.0196, +0.1457] \\
5\% & E3Sym & +0.4566 & [+0.4124, +0.5008] & [+0.3863, +0.5286] \\
5\% & Kazhdan & -0.1260 & [-0.1653, -0.0867] & [-0.1931, -0.0606] \\
5\% & Korman PAS & +0.0638 & [+0.0245, +0.1047] & [-0.0016, +0.1309] \\
5\% & Langevin & +0.1015 & [+0.0589, +0.1440] & [+0.0327, +0.1702] \\
5\% & PCA & -0.0147 & [-0.0507, +0.0213] & [-0.0736, +0.0442] \\
5\% & PRS-Net & +0.3682 & [+0.3224, +0.4141] & [+0.2913, +0.4419] \\
5\% & PRST & +0.1489 & [+0.1047, +0.1931] & [+0.0769, +0.2209] \\
5\% & SymSeg & +0.2799 & [+0.2340, +0.3257] & [+0.2062, +0.3535] \\
5\% & Hruda & +0.7054 & [+0.6678, +0.7430] & [+0.6416, +0.7643] \\
\bottomrule
\end{tabular}}\par}
\end{minipage}
\vspace{2pt}
\caption{All 40 paired top-1 retrieval comparisons, $\Delta$ = FAST minus comparator success rate. Adjustment covers the family of 40 contrasts.}\label{tab:supp_retrieval_ci}
\end{table*}

\section{Complete Component Ablations}
\label{sec:supp_ablations}

We evaluate the contribution of each detector component
on both benchmarks at all four noise levels.
Tables~\ref{tab:fast_ablations_noise00}--
\ref{tab:fast_ablations_noise05} report micro-$F_1$,
median angular error over matched true positives,
and median GPU internal detector runtime.
All variants use the same evaluation protocol and
frozen parameters except for the component being ablated.
Full FAST uses $L=48$ and level-6 projective readout.
The no-gate/no-witness variant is the ``No pair screening''
variant in the main paper.

\paragraph{Geometric voting.}
Centroid-consistency filtering and witness weighting
provide complementary mechanisms for suppressing
unreliable hypotheses. On clean inputs, removing both
reduces $F_1$ from $0.8612$ to $0.8016$ on PRS and
from $0.9466$ to $0.7931$ on \textsc{Analytic}.
Their individual contributions depend on noise level:
in particular, removing the offset gate improves PRS
$F_1$ at $3\%$ and $5\%$ noise. We analyze this
interaction and the corresponding precision--recall
tradeoff in Section~\ref{sec:supp_gate_witness}.

\paragraph{Harmonic processing and readout.}
Both raw-histogram extraction and identity harmonic
filtering reduce detection accuracy, particularly under stronger noise. The raw CIC variant also
changes peak extraction and bypasses continuous
refinement, whereas identity filtering retains harmonic
projection and synthesis but removes the Gaussian
smoothing multipliers.
Projective NMS is especially important on highly
symmetric shapes, while removing the relative peak
floor has comparatively little effect.

\paragraph{Refinement and geometric verification.}
Geometric verification is essential for rejecting
spurious hypotheses: removing it reduces clean
$F_1$ to $0.3509$ on PRS and $0.6638$ on
\textsc{Analytic}, with similarly large reductions
under noise.

Continuous refinement primarily improves angular
localization. On clean PRS, omitting refinement
increases median angular error from $0.191^\circ$
to $0.445^\circ$, while leaving $F_1$ nearly unchanged.

We also compare verification of the original hypothesis,
the refined hypothesis, or both. Full FAST verifies both
and retains the lower-residual plane for each candidate.
The original-only variant still performs refinement but
verifies only the original, distinguishing hypothesis
selection from refinement cost.

Dual verification has a mixed effect on detection $F_1$.
On \textsc{Analytic}, it improves over both single-hypothesis
variants on clean inputs and at $3\%$ and $5\%$ noise,
but original-only verification is slightly better at $1\%$.
On PRS, single-hypothesis variants achieve slightly higher
$F_1$ under noise.
Likewise, selecting the refined hypothesis alone can
yield lower angular errors than dual verification.
These results indicate that dual verification provides
an additional geometric selection option, rather than
a uniformly superior detection or localization rule.

\begin{table*}[!t]
\centering
\captionsetup{font=scriptsize,skip=0pt}
\centering
\begin{minipage}[t]{0.47\textwidth}\centering\vspace{0pt}
{\fontsize{7.4pt}{8.9pt}\selectfont\setlength{\tabcolsep}{1.6pt}\renewcommand{\arraystretch}{0.94}
\resizebox{\linewidth}{!}{\begin{tabular}{lrrrrrr}
\toprule
 & \multicolumn{3}{c}{PRS} &
   \multicolumn{3}{c}{\textsc{Analytic}} \\
\cmidrule(lr){2-4}\cmidrule(lr){5-7}
Variant & $F_1$ & Angle ($^\circ$) & ms
        & $F_1$ & Angle ($^\circ$) & ms \\
\midrule
Full FAST & 0.8612 & 0.191 & 3.44 & 0.9466 & 0.329 & 3.68 \\
\addlinespace[3pt]
\multicolumn{7}{l}{\textit{Geometric voting}} \\
No offset gate & 0.8527 & 0.178 & 3.29 & 0.9263 & 0.367 & 3.64 \\
No witnesses & 0.8593 & 0.282 & 3.18 & 0.8870 & 0.522 & 3.40 \\
No gate or witnesses & 0.8016 & 0.210 & 3.03 & 0.7931 & 0.821 & 3.38 \\
\addlinespace[3pt]
\multicolumn{7}{l}{\textit{Harmonic processing and readout}} \\
Raw CIC (mass) & 0.7241 & 0.871 & 3.84 & 0.8569 & 1.232 & 3.96 \\
Identity harmonic filter & 0.8593 & 0.250 & 5.28 & 0.9304 & 0.402 & 5.18 \\
No NMS & 0.8603 & 0.191 & 3.30 & 0.8599 & 0.318 & 3.53 \\
No relative peak floor & 0.8612 & 0.192 & 3.68 & 0.9466 & 0.329 & 3.63 \\
\addlinespace[3pt]
\multicolumn{7}{l}{\textit{Refinement and verification}} \\
No refinement & 0.8633 & 0.445 & 2.69 & 0.9436 & 0.536 & 2.92 \\
Original hypothesis only & 0.8633 & 0.445 & 3.35 & 0.9436 & 0.536 & 3.58 \\
Refined hypothesis only & 0.8615 & 0.193 & 3.35 & 0.9386 & 0.336 & 3.60 \\
No geometric verification & 0.3509 & 0.209 & 2.41 & 0.6638 & 0.336 & 2.47 \\
\bottomrule
\end{tabular}}\par}
\caption{Component ablations on clean inputs.
Angular errors are measured over matched true positives;
times are median GPU internal detector runtimes, using
the same timing boundary as the main-paper ablations.
All acceptance thresholds remain fixed.}\label{tab:fast_ablations_noise00}
\vspace{0.5em}
\end{minipage}
\hfill
\begin{minipage}[t]{0.47\textwidth}\centering\vspace{0pt}
{\fontsize{7.4pt}{8.9pt}\selectfont\setlength{\tabcolsep}{1.6pt}\renewcommand{\arraystretch}{0.94}
\resizebox{\linewidth}{!}{\begin{tabular}{lrrrrrr}
\toprule
 & \multicolumn{3}{c}{PRS} &
   \multicolumn{3}{c}{\textsc{Analytic}} \\
\cmidrule(lr){2-4}\cmidrule(lr){5-7}
Variant & $F_1$ & Angle ($^\circ$) & ms
        & $F_1$ & Angle ($^\circ$) & ms \\
\midrule
Full FAST & 0.8496 & 0.303 & 3.18 & 0.8748 & 0.496 & 3.38 \\
\addlinespace[3pt]
\multicolumn{7}{l}{\textit{Geometric voting}} \\
No offset gate & 0.8457 & 0.287 & 3.07 & 0.8536 & 0.590 & 3.23 \\
No witnesses & 0.8462 & 0.369 & 2.91 & 0.8525 & 0.711 & 3.06 \\
No gate or witnesses & 0.7854 & 0.304 & 2.84 & 0.7488 & 1.252 & 3.04 \\
\addlinespace[3pt]
\multicolumn{7}{l}{\textit{Harmonic processing and readout}} \\
Raw CIC (mass) & 0.7828 & 1.228 & 3.53 & 0.7965 & 1.702 & 3.63 \\
Identity harmonic filter & 0.8232 & 0.381 & 4.62 & 0.8540 & 0.853 & 4.60 \\
No NMS & 0.8466 & 0.303 & 3.06 & 0.7633 & 0.477 & 3.29 \\
No relative peak floor & 0.8481 & 0.305 & 3.21 & 0.8748 & 0.496 & 3.29 \\
\addlinespace[3pt]
\multicolumn{7}{l}{\textit{Refinement and verification}} \\
No refinement & 0.8533 & 0.472 & 2.46 & 0.8778 & 0.655 & 2.61 \\
Original hypothesis only & 0.8533 & 0.472 & 3.12 & 0.8778 & 0.655 & 3.28 \\
Refined hypothesis only & 0.8500 & 0.237 & 3.11 & 0.8698 & 0.463 & 3.30 \\
No geometric verification & 0.3688 & 0.253 & 2.39 & 0.6321 & 0.463 & 2.46 \\
\bottomrule
\end{tabular}}\par}
\caption{Component ablations at $1\%$ RMS noise.
Metrics and timing follow
Table~\ref{tab:fast_ablations_noise00}.}\label{tab:fast_ablations_noise01}
\end{minipage}
\end{table*}

\begin{table*}[!t]
\centering
\vspace{2pt}
\centering
\begin{minipage}[t]{0.47\textwidth}\centering\vspace{0pt}
{\fontsize{7.4pt}{8.9pt}\selectfont\setlength{\tabcolsep}{1.6pt}\renewcommand{\arraystretch}{0.94}
\resizebox{\linewidth}{!}{\begin{tabular}{lrrrrrr}
\toprule
 & \multicolumn{3}{c}{PRS} &
   \multicolumn{3}{c}{\textsc{Analytic}} \\
\cmidrule(lr){2-4}\cmidrule(lr){5-7}
Variant & $F_1$ & Angle ($^\circ$) & ms
        & $F_1$ & Angle ($^\circ$) & ms \\
\midrule
Full FAST & 0.7506 & 0.506 & 3.12 & 0.7638 & 1.121 & 3.30 \\
\addlinespace[3pt]
\multicolumn{7}{l}{\textit{Geometric voting}} \\
No offset gate & 0.7663 & 0.460 & 3.03 & 0.7520 & 1.186 & 3.21 \\
No witnesses & 0.7489 & 0.605 & 2.85 & 0.7417 & 1.368 & 3.00 \\
No gate or witnesses & 0.7106 & 0.444 & 2.79 & 0.6464 & 1.653 & 2.95 \\
\addlinespace[3pt]
\multicolumn{7}{l}{\textit{Harmonic processing and readout}} \\
Raw CIC (mass) & 0.6425 & 1.746 & 3.49 & 0.6197 & 2.504 & 3.53 \\
Identity harmonic filter & 0.6696 & 1.095 & 4.45 & 0.6865 & 2.187 & 4.43 \\
No NMS & 0.7357 & 0.505 & 3.01 & 0.6934 & 1.076 & 3.19 \\
No relative peak floor & 0.7475 & 0.506 & 3.09 & 0.7638 & 1.121 & 3.20 \\
\addlinespace[3pt]
\multicolumn{7}{l}{\textit{Refinement and verification}} \\
No refinement & 0.7551 & 0.571 & 2.41 & 0.7625 & 1.233 & 2.57 \\
Original hypothesis only & 0.7551 & 0.571 & 3.07 & 0.7625 & 1.233 & 3.22 \\
Refined hypothesis only & 0.7540 & 0.437 & 3.06 & 0.7626 & 1.107 & 3.20 \\
No geometric verification & 0.3633 & 0.441 & 2.38 & 0.5742 & 1.107 & 2.44 \\
\bottomrule
\end{tabular}}\par}
\caption{Component ablations at $3\%$ RMS noise.
Metrics and timing follow
Table~\ref{tab:fast_ablations_noise00}.}\label{tab:fast_ablations_noise03}
\vspace{0.5em}
\end{minipage}
\hfill
\begin{minipage}[t]{0.47\textwidth}\centering\vspace{0pt}
{\fontsize{7.4pt}{8.9pt}\selectfont\setlength{\tabcolsep}{1.6pt}\renewcommand{\arraystretch}{0.94}
\resizebox{\linewidth}{!}{\begin{tabular}{lrrrrrr}
\toprule
 & \multicolumn{3}{c}{PRS} &
   \multicolumn{3}{c}{\textsc{Analytic}} \\
\cmidrule(lr){2-4}\cmidrule(lr){5-7}
Variant & $F_1$ & Angle ($^\circ$) & ms
        & $F_1$ & Angle ($^\circ$) & ms \\
\midrule
Full FAST & 0.6475 & 0.699 & 3.11 & 0.6421 & 1.859 & 3.30 \\
\addlinespace[3pt]
\multicolumn{7}{l}{\textit{Geometric voting}} \\
No offset gate & 0.6786 & 0.632 & 3.03 & 0.6190 & 1.768 & 3.15 \\
No witnesses & 0.6477 & 0.871 & 2.85 & 0.6145 & 2.132 & 3.00 \\
No gate or witnesses & 0.6285 & 0.531 & 2.79 & 0.5094 & 1.948 & 2.93 \\
\addlinespace[3pt]
\multicolumn{7}{l}{\textit{Harmonic processing and readout}} \\
Raw CIC (mass) & 0.4695 & 2.306 & 3.51 & 0.4364 & 2.923 & 3.58 \\
Identity harmonic filter & 0.4969 & 1.954 & 4.48 & 0.5070 & 2.884 & 4.50 \\
No NMS & 0.6350 & 0.699 & 3.00 & 0.6130 & 1.873 & 3.17 \\
No relative peak floor & 0.6472 & 0.699 & 3.08 & 0.6421 & 1.859 & 3.20 \\
\addlinespace[3pt]
\multicolumn{7}{l}{\textit{Refinement and verification}} \\
No refinement & 0.6509 & 0.749 & 2.41 & 0.6257 & 1.875 & 2.55 \\
Original hypothesis only & 0.6509 & 0.749 & 3.07 & 0.6257 & 1.875 & 3.21 \\
Refined hypothesis only & 0.6552 & 0.651 & 3.06 & 0.6296 & 1.795 & 3.20 \\
No geometric verification & 0.3310 & 0.664 & 2.38 & 0.4782 & 1.835 & 2.44 \\
\bottomrule
\end{tabular}}\par}
\caption{Component ablations at $5\%$ RMS noise.
Metrics and timing follow
Table~\ref{tab:fast_ablations_noise00}.}\label{tab:fast_ablations_noise05}
\end{minipage}
\end{table*}

\subsection{Bandlimit and Readout Resolution}
\label{sec:supp_resolution}

Table~\ref{tab:supp_resolution} separates the effects of harmonic
bandlimit and directional readout density. We vary $L$ at fixed
level-6 readout, and vary the icosphere level at fixed $L=48$.
Changing $L$ also changes the histogram to $(L+2)\times(4L+1)$
and the multiplier quadrature to $\max(32,4L)$ nodes; the kernel
bandwidth, thresholds, and other detector settings remain fixed.
All accuracy entries use the complete benchmarks at each noise level.
The clean times are GPU internal detector medians, matching the
component-ablation timing boundary.

Lowering $L$ to 16 degrades
clean Analytic $F_1$ from $0.947$ to $0.797$.
At $L=48$, increasing readout from level 4 to level 6 improves clean
Analytic $F_1$ from $0.914$ to $0.947$ and reduces PRS angular
error at $5\%$ noise from $1.034^\circ$ to $0.699^\circ$.
The effects are not monotonic in every metric: level 5, for example,
has higher Analytic $F_1$ than level 6 at $1\%$ and $3\%$ noise.

\begin{table*}[t]
\centering
{\fontsize{7.8pt}{9.2pt}\selectfont
\setlength{\tabcolsep}{3pt}\renewcommand{\arraystretch}{1.02}
\begin{tabular}{llrrrrrrrrr}
\toprule
 & & \multicolumn{3}{c}{Clean} & \multicolumn{2}{c}{1\%}
 & \multicolumn{2}{c}{3\%} & \multicolumn{2}{c}{5\%} \\
Dataset & Setting & $F_1$ & Angle & ms & $F_1$ & Angle
 & $F_1$ & Angle & $F_1$ & Angle \\
\midrule
\multicolumn{11}{l}{\textit{Bandlimit sweep at level 6 ($N_u=20{,}481$)}} \\
PRS & $L=16$ & 0.854 & 0.389 & 3.91 & 0.843 & 0.430 & 0.765 & 0.559 & 0.663 & 0.721 \\
PRS & $L=32$ & 0.860 & 0.223 & 3.42 & 0.846 & 0.313 & 0.752 & 0.518 & 0.648 & 0.720 \\
PRS & $L=48$ & 0.861 & 0.191 & 3.44 & 0.850 & 0.303 & 0.751 & 0.506 & 0.647 & 0.699 \\
PRS & $L=64$ & 0.861 & 0.185 & 3.48 & 0.850 & 0.297 & 0.749 & 0.502 & 0.643 & 0.694 \\
PRS & $L=80$ & 0.863 & 0.184 & 3.55 & 0.851 & 0.295 & 0.748 & 0.497 & 0.643 & 0.696 \\
PRS & $L=96$ & 0.863 & 0.183 & 3.70 & 0.851 & 0.295 & 0.749 & 0.501 & 0.642 & 0.683 \\
\addlinespace[2pt]
\textsc{Analytic} & $L=16$ & 0.797 & 0.472 & 3.99 & 0.776 & 0.584 & 0.719 & 1.108 & 0.611 & 1.580 \\
\textsc{Analytic} & $L=32$ & 0.937 & 0.348 & 3.68 & 0.870 & 0.517 & 0.759 & 1.099 & 0.638 & 1.847 \\
\textsc{Analytic} & $L=48$ & 0.947 & 0.329 & 3.68 & 0.875 & 0.496 & 0.764 & 1.121 & 0.642 & 1.859 \\
\textsc{Analytic} & $L=64$ & 0.944 & 0.324 & 3.87 & 0.879 & 0.499 & 0.771 & 1.121 & 0.637 & 1.884 \\
\textsc{Analytic} & $L=80$ & 0.944 & 0.324 & 3.87 & 0.881 & 0.500 & 0.770 & 1.119 & 0.637 & 1.837 \\
\textsc{Analytic} & $L=96$ & 0.944 & 0.319 & 3.87 & 0.885 & 0.497 & 0.770 & 1.099 & 0.630 & 1.873 \\
\midrule
\multicolumn{11}{l}{\textit{Readout sweep at $L=48$}} \\
PRS & Level 4 & 0.860 & 0.262 & 3.36 & 0.845 & 0.316 & 0.744 & 0.577 & 0.644 & 1.034 \\
PRS & Level 5 & 0.864 & 0.193 & 3.36 & 0.848 & 0.280 & 0.748 & 0.608 & 0.645 & 0.877 \\
PRS & Level 6 & 0.861 & 0.191 & 3.44 & 0.850 & 0.303 & 0.751 & 0.506 & 0.647 & 0.699 \\
\addlinespace[2pt]
\textsc{Analytic} & Level 4 & 0.914 & 0.385 & 3.65 & 0.861 & 0.541 & 0.745 & 1.344 & 0.638 & 1.934 \\
\textsc{Analytic} & Level 5 & 0.940 & 0.336 & 3.59 & 0.891 & 0.596 & 0.777 & 1.257 & 0.641 & 1.852 \\
\textsc{Analytic} & Level 6 & 0.947 & 0.329 & 3.68 & 0.875 & 0.496 & 0.764 & 1.121 & 0.642 & 1.859 \\
\bottomrule
\end{tabular}}
\caption{Bandlimit and readout-resolution ablations. Angle denotes
median matched-true-positive angular error in degrees; runtimes are
clean-input median GPU internal detector times in milliseconds.
Levels 4, 5, and 6 contain 1281, 5121, and 20481 projective
directions, respectively. Full FAST uses $L=48$ and level 6.}
\label{tab:supp_resolution}
\end{table*}

\section{Gate--Witness Interaction under Noise}
\label{sec:supp_gate_witness}

We evaluate a complete $2\times2$ factorial design, independently
enabling or disabling the centroid-consistency gate and witness
weighting. All four variants retain the same harmonic kernel,
readout, verifier, and frozen thresholds across both benchmarks
and all four noise levels. Table~\ref{tab:supp_gate_witness_factorial}
reports precision, recall, and micro-$F_1$ for all $32$ dataset--noise--configuration combinations.

\begin{table*}[!t]
\centering
\vspace{2pt}
\begin{minipage}[t]{0.47\textwidth}\centering\vspace{0pt}
{\fontsize{7.7pt}{8.8pt}\selectfont\setlength{\tabcolsep}{2pt}\renewcommand{\arraystretch}{0.96}
\resizebox{\linewidth}{!}{\begin{tabular}{@{}lllcc@{}}
\toprule
Noise & Gate & Witness & PRS: $P/R/F_1$ & Analytic: $P/R/F_1$ \\
\midrule
0\% & On & On & 0.839 / 0.885 / 0.861 & 0.941 / 0.952 / 0.947 \\
0\% & Off & On & 0.838 / 0.868 / 0.853 & 0.904 / 0.949 / 0.926 \\
0\% & On & Off & 0.841 / 0.878 / 0.859 & 0.892 / 0.882 / 0.887 \\
0\% & Off & Off & 0.843 / 0.764 / 0.802 & 0.821 / 0.767 / 0.793 \\
1\% & On & On & 0.776 / 0.938 / 0.850 & 0.868 / 0.881 / 0.875 \\
1\% & Off & On & 0.783 / 0.919 / 0.846 & 0.834 / 0.874 / 0.854 \\
1\% & On & Off & 0.781 / 0.923 / 0.846 & 0.859 / 0.846 / 0.852 \\
1\% & Off & Off & 0.802 / 0.770 / 0.785 & 0.766 / 0.733 / 0.749 \\
3\% & On & On & 0.621 / 0.947 / 0.751 & 0.785 / 0.744 / 0.764 \\
3\% & Off & On & 0.652 / 0.930 / 0.766 & 0.754 / 0.750 / 0.752 \\
3\% & On & Off & 0.630 / 0.924 / 0.749 & 0.775 / 0.711 / 0.742 \\
3\% & Off & Off & 0.715 / 0.706 / 0.711 & 0.697 / 0.603 / 0.646 \\
5\% & On & On & 0.504 / 0.906 / 0.647 & 0.679 / 0.609 / 0.642 \\
5\% & Off & On & 0.550 / 0.887 / 0.679 & 0.643 / 0.597 / 0.619 \\
5\% & On & Off & 0.512 / 0.880 / 0.648 & 0.653 / 0.581 / 0.614 \\
5\% & Off & Off & 0.638 / 0.620 / 0.628 & 0.562 / 0.466 / 0.509 \\
\bottomrule
\end{tabular}}\par}
\caption{Gate--witness factorial study across noise levels. All four configurations use the same remaining detector settings and acceptance threshold.}\label{tab:supp_gate_witness_factorial}
\end{minipage}
\hfill
\begin{minipage}[t]{0.47\textwidth}\centering\vspace{0pt}
{\fontsize{7.7pt}{8.8pt}\selectfont\setlength{\tabcolsep}{2pt}\renewcommand{\arraystretch}{0.96}
\resizebox{\linewidth}{!}{\begin{tabular}{@{}llcrl@{}}
\toprule
Dataset & Noise & Metric & $\Delta$ & Adjusted 95\% CI \\
\midrule
PRS & 0\% & $F_1$ & +0.0595 & [+0.0384, +0.0820] \\
 &  & $R$ & +0.1208 & [+0.0901, +0.1531] \\
PRS & 1\% & $F_1$ & +0.0642 & [+0.0393, +0.0902] \\
 &  & $R$ & +0.1689 & [+0.1336, +0.2059] \\
PRS & 3\% & $F_1$ & +0.0400 & [+0.0041, +0.0761] \\
 &  & $R$ & +0.2416 & [+0.2020, +0.2816] \\
PRS & 5\% & $F_1$ & +0.0190 & [-0.0204, +0.0591] \\
 &  & $R$ & +0.2864 & [+0.2441, +0.3286] \\
\textsc{Analytic} & 0\% & $F_1$ & +0.1535 & [+0.1029, +0.2125] \\
 &  & $R$ & +0.1854 & [+0.1137, +0.2678] \\
\textsc{Analytic} & 1\% & $F_1$ & +0.1260 & [+0.0788, +0.1810] \\
 &  & $R$ & +0.1489 & [+0.0824, +0.2241] \\
\textsc{Analytic} & 3\% & $F_1$ & +0.1174 & [+0.0589, +0.1812] \\
 &  & $R$ & +0.1408 & [+0.0725, +0.2107] \\
\textsc{Analytic} & 5\% & $F_1$ & +0.1327 & [+0.0746, +0.1872] \\
 &  & $R$ & +0.1429 & [+0.0830, +0.1996] \\
\bottomrule
\end{tabular}}\par}
\caption{Paired full-FAST minus no-gate/no-witness effects for F1 and recall. Bonferroni adjustment covers 16 prespecified contrasts (two datasets, four noise levels, two metrics).}\label{tab:supp_gate_witness_joint_ci}
\end{minipage}
\end{table*}

\paragraph{Joint contribution.}
Table~\ref{tab:supp_gate_witness_joint_ci} reports paired full-minus-neither
$F_1$ and recall differences, adjusted across $16$ comparisons. Removing
both components reduces measured $F_1$ at every noise level; the adjusted
advantage is supported in seven of eight dataset--noise conditions, except
PRS at $5\%$. Recall improves significantly in all eight conditions.

\paragraph{Interaction at higher noise.}
With witnesses retained, removing the gate on PRS at $3\%$ noise raises
precision from $0.621$ to $0.652$ and lowers recall from $0.947$ to
$0.930$, increasing $F_1$ from $0.751$ to $0.766$. At $5\%$ noise,
precision rises from $0.504$ to $0.550$, recall falls from $0.906$
to $0.887$, and $F_1$ rises from $0.647$ to $0.679$.
These gate-only differences are descriptive point estimates;
the adjusted intervals in Table~\ref{tab:supp_gate_witness_joint_ci}
refer to the separate full-minus-neither contrast. Thus, the
gate--witness interaction is noise-dependent, not uniformly beneficial.

\clearpage
\endgroup

\end{document}